\documentclass{article}
\PassOptionsToPackage{numbers, compress}{natbib}

\usepackage[final]{neurips_2020}

\usepackage[utf8]{inputenc} \usepackage[T1]{fontenc}    \usepackage{hyperref}       \usepackage{url}            \usepackage{booktabs}       \usepackage{amsfonts}       \usepackage{nicefrac}       \usepackage{xcolor}
\usepackage{epsfig}
\usepackage{algorithm}
\usepackage{algorithmic}
\usepackage{amsthm}
\usepackage{amsmath}
\usepackage{microtype}
\usepackage{mathtools}
\usepackage{float}
\usepackage{graphicx}
\usepackage{wrapfig}
\usepackage{makecell}
\usepackage{caption}
\usepackage{subcaption}
\usepackage{bm}
\usepackage{bbm}
\usepackage{multirow}
\usepackage{microtype}
\usepackage{mathtools}
\usepackage{enumitem}
\usepackage{cancel}
\usepackage{chngcntr}

\newcommand{\one}[1]{\mathbbm{1}_{[#1]}}

\title{Big Self-Supervised Models are\\Strong Semi-Supervised Learners}

\author{
Ting Chen\thanks{Correspondence to: \texttt{iamtingchen@google.com}}, ~Simon Kornblith, ~Kevin Swersky, ~Mohammad Norouzi, ~Geoffrey Hinton\\
Google Research, Brain Team\\
}
\renewcommand\footnotemark{}

\begin{document}
\maketitle

\begin{abstract}
One paradigm for learning from few labeled examples while making best use of a large amount of unlabeled data is \textit{unsupervised pretraining} followed by \textit{supervised fine-tuning}. 
Although this paradigm uses unlabeled data in a \textit{task-agnostic} way, in contrast to common approaches to semi-supervised learning for computer vision, we show that it is surprisingly effective for semi-supervised learning on ImageNet.
A key ingredient of our approach is the use of big (deep and wide) networks during pretraining and fine-tuning. We find that, the fewer the labels, the more this approach (task-agnostic use of unlabeled data) benefits from a bigger network. After fine-tuning, the big network can be further improved and distilled into a much smaller one with little loss in classification accuracy by using the unlabeled examples for a second time, but in a \textit{task-specific} way.
The proposed semi-supervised learning algorithm can be summarized in three steps: \textit{unsupervised pretraining} of a big ResNet model using SimCLRv2, \textit{supervised fine-tuning} on a few labeled examples, and \textit{distillation with unlabeled examples} for refining and transferring the task-specific knowledge. This procedure achieves 73.9\% ImageNet top-1 accuracy with just 1\% of the labels ($\le$13 labeled images per class) using ResNet-50, a $10\times$ improvement in label efficiency over the previous state-of-the-art. With 10\% of labels, ResNet-50 trained with our method achieves 77.5\% top-1 accuracy, outperforming standard supervised training with all of the labels. \footnote{Code and pretrained checkpoints are available at \href{https://github.com/google-research/simclr}{https://github.com/google-research/simclr}.}
\end{abstract} \section{Introduction}

\begin{figure}[!h]
\begin{minipage}[c]{0.3\linewidth}
    \centering
    \small
    \includegraphics[width=0.92\linewidth]{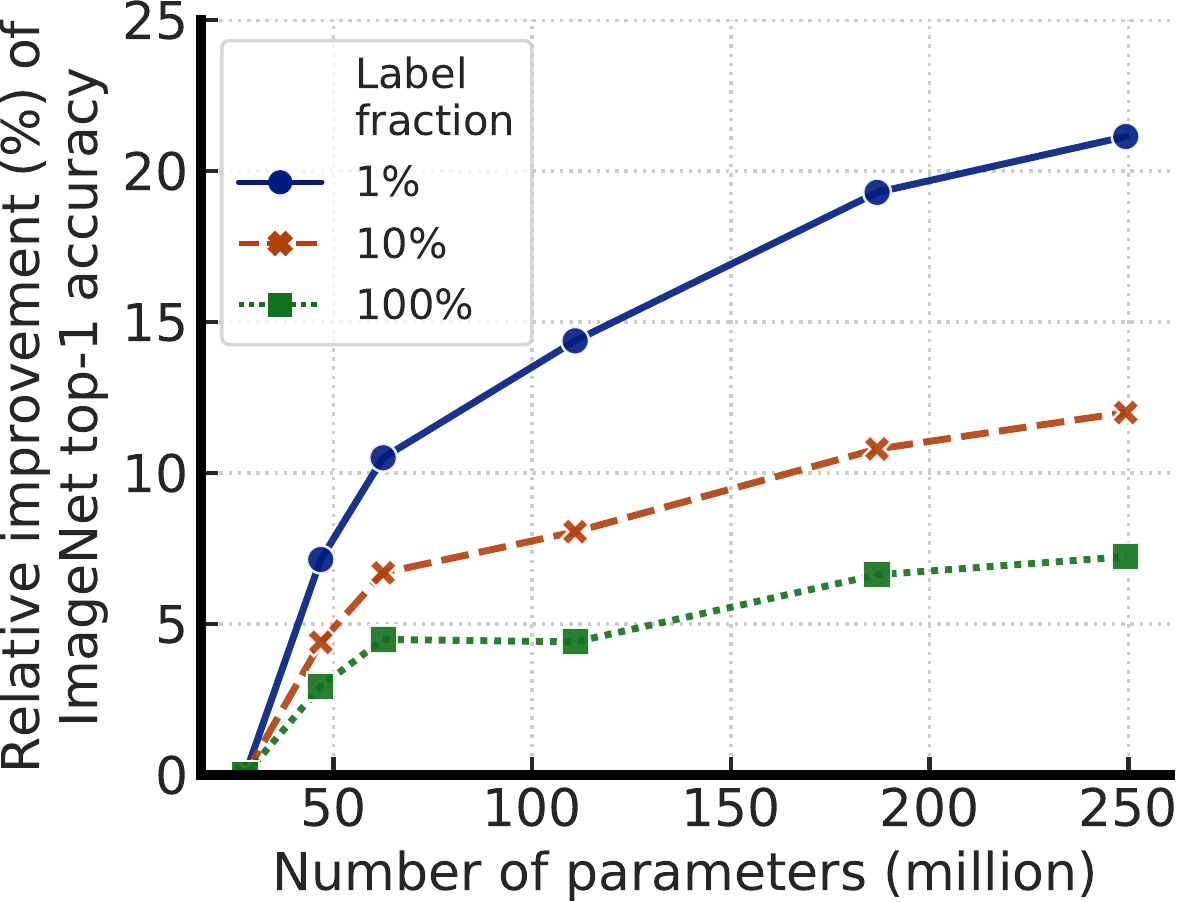}
    \caption{\label{fig:semi_param} Bigger models yield larger gains when fine-tuning with fewer labeled examples.}
\end{minipage}\hfill
\begin{minipage}[c]{0.68\linewidth}
    \centering
    \small
    \begin{subfigure}{.44\textwidth}
      \centering
\includegraphics[width=0.98\linewidth]{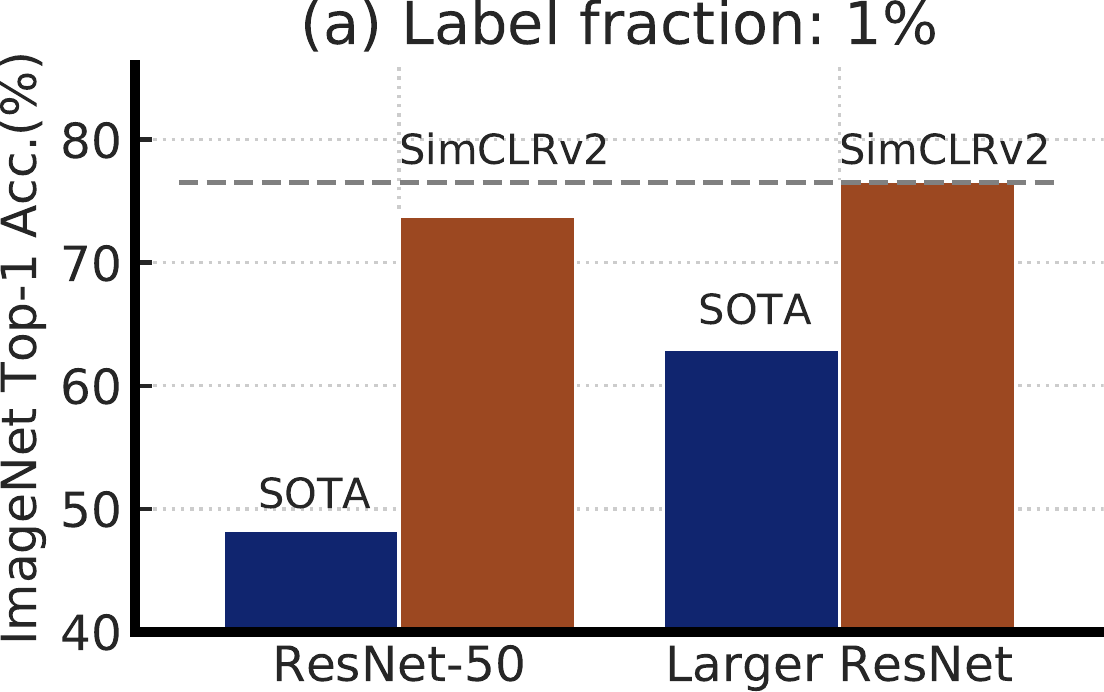}
    \end{subfigure}
    \begin{subfigure}{.55\textwidth}
      \centering
\includegraphics[width=0.98\linewidth]{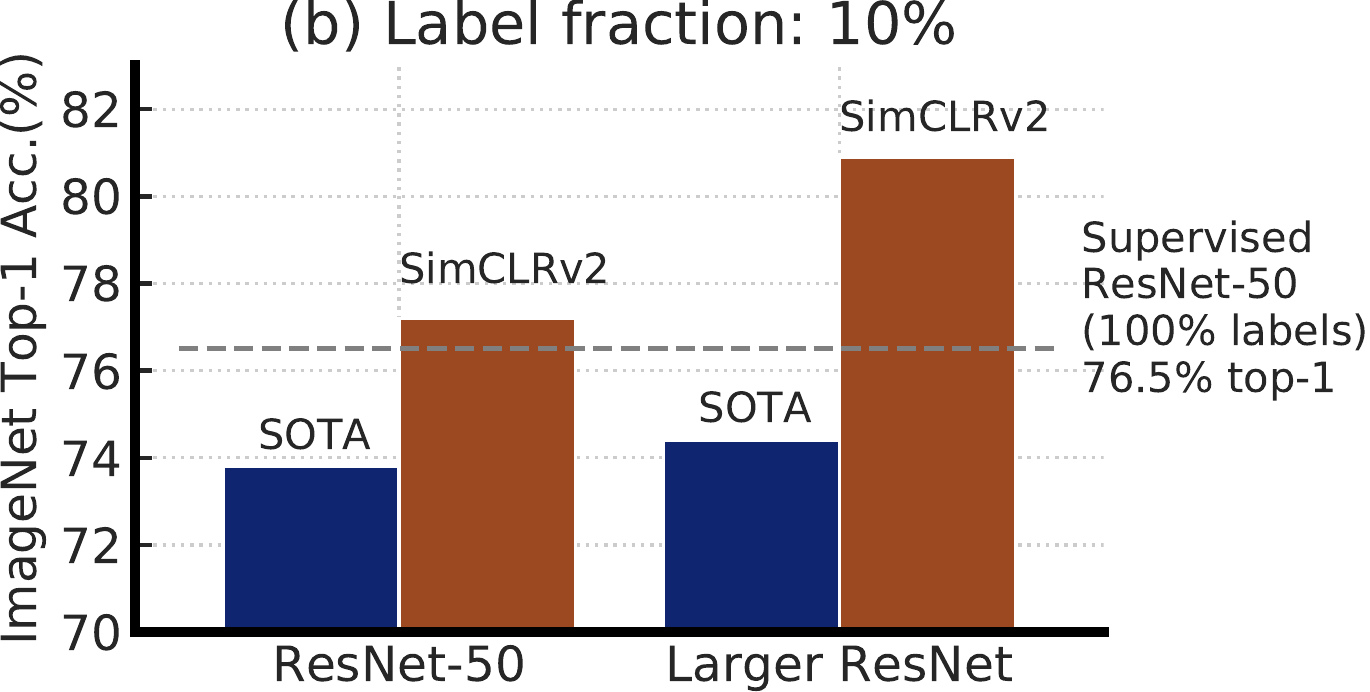}
    \end{subfigure}
    \caption{\label{fig:sota_cp} Top-1 accuracy of previous state-of-the-art (SOTA) methods~\cite{chen2020simple,pham2020meta} and our method (SimCLRv2) on ImageNet using only 1\% or 10\% of the labels. Dashed line denotes fully supervised ResNet-50 trained with 100\% of labels. Full comparisons in Table ~\ref{tab:sota_more}.}
\end{minipage}\hfill
\end{figure}
Learning from just a few labeled examples while making best use of a large amount of unlabeled data is a long-standing problem in machine learning.
One approach to semi-supervised learning involves unsupervised or self-supervised pretraining, followed by supervised fine-tuning~\cite{hinton2006fast,bengio2007greedy}.
This approach leverages unlabeled data in a \textit{task-agnostic} way during pretraining, as the supervised labels are only used during fine-tuning. Although it has received little attention in computer vision, this approach has become predominant in natural language processing, where one first trains a large language model on unlabeled text (e.g., Wikipedia),
and then fine-tunes the model on a few labeled examples~\cite{dai2015semi,kiros2015skip,radford2018improving,peters2018deep,devlin2018bert,radford2019language}.
An alternative approach, common in computer vision, directly leverages unlabeled data during supervised learning, as a form of regularization.
This approach uses unlabeled data in a \textit{task-specific} way to encourage class label prediction consistency on unlabeled data among different models~\cite{lee2013pseudo,xie2019self,pham2020meta} or under different data augmentations~\cite{berthelot2019mixmatch,xie2019unsupervised,sohn2020fixmatch}.

Motivated by recent advances in self-supervised learning of visual representations~\cite{oord2018representation,wu2018unsupervised,bachman2019learning,henaff2019data,he2019momentum,chen2020simple}, this paper first presents a thorough investigation of the ``unsupervised pretrain, supervised fine-tune'' paradigm for semi-supervised learning on ImageNet~\cite{russakovsky2015imagenet}.
During self-supervised pretraining, images are used without class labels (in a task-agnostic way), hence the representations are not directly tailored to a specific classification task. With this task-agnostic use of unlabeled data, we find that network size is important: Using a big (deep and wide) neural network for self-supervised pretraining and fine-tuning greatly improves accuracy. In addition to the network size, we characterize a few important design choices for contrastive representation learning that benefit supervised fine-tuning and semi-supervised learning.

Once a convolutional network is pretrained and fine-tuned, we find that its task-specific predictions can be further improved and distilled into a smaller network. To this end, we make use of unlabeled data for a second time to encourage the student network to mimic the teacher network's label predictions. Thus, the distillation~\cite{hinton2015distilling,bucilua2006model} phase of our method using unlabeled data is reminiscent of the use of pseudo labels~\cite{lee2013pseudo} in self-training~\cite{yalniz2019billion,xie2019self}, but without much extra complexity.

\begin{figure}
    \centering
    \includegraphics[width=0.8\textwidth]{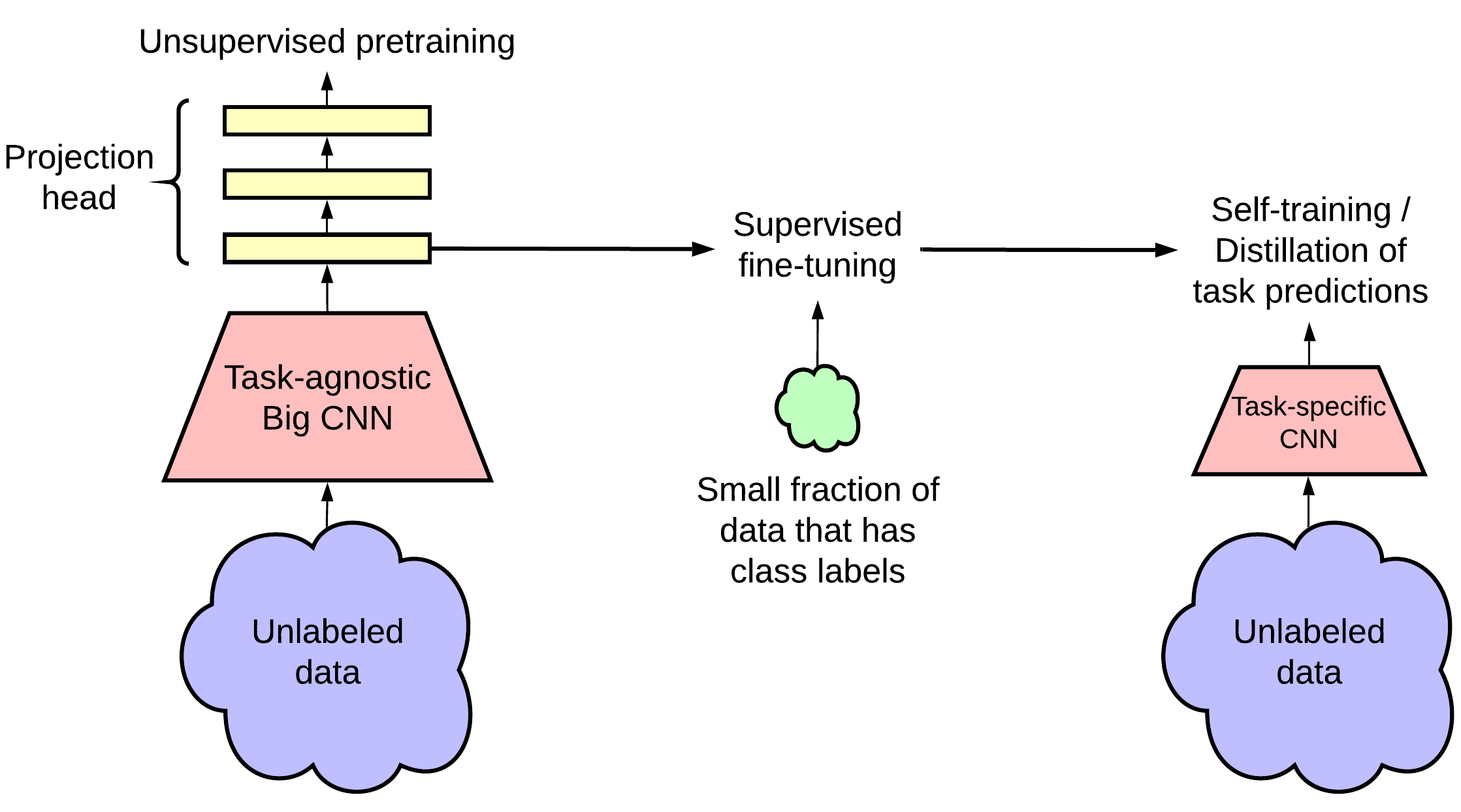}
    \caption{\label{fig:framework}The proposed semi-supervised learning framework leverages unlabeled data in two ways: (1)~task-agnostic use in unsupervised pretraining, and (2) task-specific use in self-training / distillation.}
\end{figure}

In summary, the proposed semi-supervised learning framework comprises three steps as shown in Figure~\ref{fig:framework}: (1) unsupervised or self-supervised pretraining, (2) supervised fine-tuning, and 
(3)~distillation using unlabeled data.
We develop an improved variant of a recently proposed contrastive learning framework, SimCLR~\cite{chen2020simple}, for unsupervised pretraining of a ResNet architecture~\cite{he2016deep}.
We call this framework SimCLRv2.
We assess the effectiveness of our method on ImageNet ILSVRC-2012~\cite{russakovsky2015imagenet} with only 1\% and 10\% of the labeled images available. Our main findings and contributions can be summarized as follows:
\begin{itemize}[topsep=0pt, partopsep=0pt, leftmargin=15pt, parsep=0pt, itemsep=5pt]
    \item Our empirical results suggest that for semi-supervised learning (via the task-agnostic use of unlabeled data), the fewer the labels, the more it is possible to benefit from a bigger model (Figure~\ref{fig:semi_param}). Bigger self-supervised models are more label efficient, performing significantly better when fine-tuned on only a few labeled examples, even though they have more capacity to potentially overfit.
    \item We show that although big models are important for learning general (visual) representations, the extra capacity may not be necessary when a specific target task is concerned. Therefore, with the task-specific use of unlabeled data, the predictive performance of the model can be further improved and transferred into a smaller network.
\item We further demonstrate the importance of the nonlinear transformation (a.k.a. projection head) after convolutional layers used in SimCLR for semi-supervised learning. A deeper projection head not only improves the representation quality measured by linear evaluation, but also improves semi-supervised performance when fine-tuning from a \textit{middle layer} of the projection head.
\end{itemize}

We combine these findings to achieve a new state-of-the-art in semi-supervised learning on ImageNet as summarized in Figure~\ref{fig:sota_cp}. Under the linear evaluation protocol, SimCLRv2 achieves 79.8\% top-1 accuracy, a 4.3\% relative improvement over the previous state-of-the-art~\cite{chen2020simple}.
When fine-tuned on only 1\% / 10\% of labeled examples and distilled to the same architecture using unlabeled examples, it achieves 76.6\% / 80.9\% top-1 accuracy, which is a 21.6\% / 8.7\% relative improvement over previous state-of-the-art. With distillation, these improvements can also be transferred to smaller ResNet-50 networks to achieve 73.9\% / 77.5\% top-1 accuracy using 1\% / 10\% of labels. By comparison, a standard supervised ResNet-50 trained on all of labeled images achieves a top-1 accuracy of 76.6\%.

 \section{Method}

Inspired by the recent successes of learning from unlabeled data~\cite{henaff2019data,he2019momentum,chen2020simple,lee2013pseudo,yalniz2019billion,xie2019self}, the proposed semi-supervised learning framework leverages unlabeled data in both task-agnostic and task-specific ways. The first time the unlabeled data is used, it is in a task-agnostic way, for learning general (visual) representations via unsupervised pretraining. The general representations are then adapted for a specific task via supervised fine-tuning. The second time the unlabeled data is used, it is in a task-specific way, for further improving predictive performance and obtaining a compact model. To this end, we train student networks on the unlabeled data with imputed labels from the fine-tuned teacher network. Our method can be summarized in three main steps: \textit{pretrain}, \textit{fine-tune}, and then \textit{distill}. The procedure is illustrated in Figure~\ref{fig:framework}. We introduce each specific component in detail below.

\textbf{Self-supervised pretraining with SimCLRv2.} To learn general visual representations effectively with unlabeled images, we adopt and improve SimCLR~\cite{chen2020simple}, a recently proposed approach based on contrastive learning. SimCLR learns representations by maximizing agreement~\cite{becker1992self} between differently augmented views of the same data example via a contrastive loss in the latent space. More specifically, given a randomly sampled mini-batch of images, each image $\bm x_i$ is augmented twice using random crop, color distortion and Gaussian blur, creating two views of the same example $\bm x_{2k-1}$ and $\bm x_{2k}$. The two images are encoded via an encoder network $f(\cdot)$ (a ResNet~\cite{he2016deep}) to generate representations $\bm h_{2k-1}$ and $\bm h_{2k}$. The representations are then transformed again with a non-linear transformation network $g(\cdot)$ (a MLP projection head), yielding $\bm z_{2k-1}$ and $\bm z_{2k}$ that are used for the contrastive loss. With a mini-batch of augmented examples, the contrastive loss between a pair of positive example $i, j$ (augmented from the same image) is given as follows:
\begin{equation}
\label{eq:nt_xent}
    \ell^{\mathrm{NT}\text{-}\mathrm{Xent}}_{i,j} = -\log \frac{\exp(\mathrm{sim}(\bm z_i, \bm z_j)/\tau)}{\sum_{k=1}^{2N} \one{k \neq i}\exp(\mathrm{sim}(\bm z_i, \bm z_k)/\tau)}~,
\end{equation}
Where $\mathrm{sim}(\cdot,\cdot)$ is cosine similarity between two vectors, and $\tau$ is a temperature scalar.

In this work, we propose SimCLRv2, which improves upon SimCLR~\cite{chen2020simple} in three major ways. Below we summarize the changes as well as their improvements of accuracy on Imagenet ILSVRC-2012~\cite{russakovsky2015imagenet}.
\begin{itemize}[topsep=0pt, partopsep=0pt, leftmargin=13pt, parsep=0pt, itemsep=4pt]
    \item[1.] To fully leverage the power of general pretraining, we explore larger ResNet models. Unlike SimCLR~\cite{chen2020simple} and other previous work~\cite{kolesnikov2019revisiting,he2019momentum}, whose largest model is ResNet-50 (4$\times$), we train models that are deeper but less wide. The largest model we train is a 152-layer ResNet~\cite{he2016deep} with 3$\times$ wider channels and selective kernels (SK)~\cite{li2019selective}, a channel-wise attention mechanism that improves the parameter efficiency of the network. 
    By scaling up the model from ResNet-50 to ResNet-152 ($3\times$+SK), we obtain a 29\% relative improvement in top-1 accuracy when fine-tuned on 1\% of labeled examples.
    \item[2.] We also increase the capacity of the non-linear network $g(\cdot)$ (a.k.a. projection head), by making it deeper.\footnote{In our experiments, we set the width of projection head's middle layers to that of its input, so it is also adjusted by the width multiplier. However, a wider projection head improves performance even when the base network remains narrow.} Furthermore, instead of throwing away $g(\cdot)$ entirely after pretraining as in SimCLR~\cite{chen2020simple}, we fine-tune from a middle layer (detailed later).
This small change yields a significant improvement for both linear evaluation and fine-tuning with only a few labeled examples.
    Compared to SimCLR with 2-layer projection head, by using a 3-layer projection head and fine-tuning from the 1st layer of projection head, it results in as much as 14\% relative improvement in top-1 accuracy when fine-tuned on 1\% of labeled examples (see Figure~\ref{fig:param_head}).
    \item[3.] Motivated by~\cite{chen2020improved}, we also incorporate the memory mechanism from MoCo~\cite{he2019momentum}, which designates a memory network (with a moving average of weights for stabilization) whose output will be buffered as negative examples. Since our training is based on large mini-batch which already supplies many contrasting negative examples, this change yields an improvement of $\sim$1\% for linear evaluation as well as when fine-tuning on 1\% of labeled examples (see Appendix~\ref{app:moco}).
\end{itemize}

\textbf{Fine-tuning.} 
Fine-tuning is a common way to adapt the task-agnostically pretrained network for a specific task.
In SimCLR~\cite{chen2020simple}, the MLP projection head $g(\cdot)$ is discarded entirely after pretraining, while only the ResNet encoder $f(\cdot)$ is used during the fine-tuning. Instead of throwing it all away, we propose to incorporate part of the MLP projection head into the base encoder during the fine-tuning. In other words, we fine-tune the model from a \textit{middle layer} of the projection head, instead of the input layer of the projection head as in SimCLR.
Note that fine-tuning from the first layer of the MLP head is the same as adding an fully-connected layer to the base network and removing an fully-connected layer from the head, and the impact of this extra layer is contingent on the amount of labeled examples during fine-tuning (as shown in our experiments).

\textbf{Self-training / knowledge distillation via unlabeled examples.} To further improve the network for the target task, here we leverage the unlabeled data directly for the target task. Inspired by~\cite{bucilua2006model,lee2013pseudo,hinton2015distilling,yalniz2019billion,xie2019self}, we use the fine-tuned network as a teacher to impute labels for training a student network. Specifically, we minimize the following distillation loss where no real labels are used:
\begin{equation}
\label{eq:distill}
    \mathcal{L}^{\mathrm{distill}} = - \sum_{\bm x_i\in \mathcal{D}}\bigg[\sum_y P^T(y | \bm x_i;\tau) \log P^S(y|\bm x_i;\tau)\bigg]
\end{equation}
where $P(y|\bm x_i) = \exp(f^{\mathrm{task}}(\bm x_i)[y] / \tau)/ \sum_{y'}\exp(f^{\mathrm{task}}(\bm x_i)[y'] / \tau)$, and $\tau$ is a scalar temperature parameter. The teacher network, which produces $P^T(y|\bm x_i)$, is fixed during the distillation; only the student network, which produces $P^S(y|\bm x_i)$, is trained. 

While we focus on distillation using only unlabeled examples in this work, when the number of labeled examples is significant, one can also combine the distillation loss with ground-truth labeled examples using a weighted combination
\begin{equation}
    \label{eq:distill_plus}
    \mathcal{L} = -(1-\alpha)\sum_{(\bm x_i, y_i)\in \mathcal{D}^L}\bigg[\log P^S(y_i|\bm x_i) \bigg] - \alpha \sum_{\bm x_i\in \mathcal{D}}\bigg[\sum_y P^T(y | \bm x_i;\tau) \log P^S(y|\bm x_i;\tau)\bigg].
\end{equation}
This procedure can be performed using students either with the same model architecture (self-distillation), which further improves the task-specific performance, or with a smaller model architecture, which leads to a compact model.
 \section{Empirical Study}

\subsection{Settings and Implementation Details}

Following the semi-supervised learning setting in~\cite{zhai2019s4l,henaff2019data,chen2020simple}, we evaluate the proposed method on ImageNet ILSVRC-2012~\cite{russakovsky2015imagenet}. While all $\sim$1.28 million images are available, only a randomly sub-sampled 1\% (12811) or 10\% (128116) of images are associated with labels.\footnote{See \href{https://www.tensorflow.org/datasets/catalog/imagenet2012\_subset}{https://www.tensorflow.org/datasets/catalog/imagenet2012\_subset} for the details of the 1\%/10\% subsets.} As in previous work, we also report performance when training a linear classifier on top of a fixed representation with all labels~\cite{zhang2016colorful,oord2018representation,wu2018unsupervised,chen2020simple} to directly evaluate SimCLRv2 representations. We use the LARS optimizer~\cite{you2017large} (with a momentum of 0.9) throughout for pretraining, fine-tuning and distillation.

For pretraining, similar to~\cite{chen2020simple}, we train our model on 128 Cloud TPUs, with a batch size of 4096 and global batch normalization~\cite{ioffe2015batch}, for total of 800 epochs. The learning rate is linearly increased for the first 5\% of epochs, reaching maximum of 6.4 ($=0.1\times \mathrm{sqrt(BatchSize)}$), and then decayed with a cosine decay schedule.
A weight decay of $1e^{-4}$ is used. We use a 3-layer MLP projection head on top of a ResNet encoder. The memory buffer is set to 64K, and exponential moving average (EMA) decay is set to 0.999 according to~\cite{he2019momentum}. We use the same set of simple augmentations as SimCLR~\cite{chen2020simple}, namely random crop, color distortion, and Gaussian blur.

For fine-tuning, by default we fine-tune from the first layer of the projection head for 1\%/10\% of labeled examples, but from the input of the projection head when 100\% labels are present. We use global batch normalization, but we remove weight decay, learning rate warmup, and use a much smaller learning rate, i.e. 0.16 ($=0.005\times \mathrm{sqrt(BatchSize)}$) for standard ResNets~\cite{he2016deep}, and 0.064 ($=0.002\times \mathrm{sqrt(BatchSize)}$) for larger ResNets variants (with width multiplier larger than 1 and/or SK~\cite{li2019selective}). A batch size of 1024 is used. Similar to~\cite{chen2020simple}, we fine-tune for 60 epochs with 1\% of labels, and 30 epochs with 10\% of labels, as well as full ImageNet labels.

For distillation, we only use unlabeled examples, unless otherwise specified. 
We consider two types of distillation: self-distillation where the student has the same model architecture as the teacher (excluding projection head), and big-to-small distillation where the student is a much smaller network. We set temperature to 0.1 for self-distillation, and 1.0 for large-to-small distillation (though the effect of temperatures between 0.1 and 1 is very small).
We use the same learning rate schedule, weight decay, batch size as pretraining, and the models are trained for 400 epochs. Only random crop and horizontal flips of training images are applied during fine-tuning and distillation.

\subsection{Bigger Models Are More Label-Efficient}

\begin{table}[]
\small
\centering
\caption{Top-1 accuracy of fine-tuning SimCLRv2 models (on varied label fractions) or training a linear classifier on the representations. The supervised baselines are trained from scratch using all labels in 90 epochs. The parameter count only include ResNet up to final average pooling layer. For fine-tuning results with 1\% and 10\% labeled examples, the models include additional non-linear projection layers, which incurs additional parameter count (4M for $1\times$ models, and 17M for $2\times$ models). See Table~\ref{tab:big_ft_top5} for Top-5 accuracy.}
\label{tab:big_ft_top1}
\begin{tabular}{c@{\hspace{0.35cm}}cc@{\hspace{0.35cm}}c@{\hspace{0.35cm}}r@{\hspace{0.35cm}}r@{\hspace{0.35cm}}r@{\hspace{0.35cm}}r@{\hspace{0.35cm}}r}
\toprule
\multirow{2}{*}{Depth}& \multirow{2}{*}{Width}& \multirow{2}{*}{Use SK~\cite{li2019selective}}& \multirow{2}{*}{Param (M)}& \multicolumn{3}{c}{Fine-tuned on} & \multirow{2}{*}{Linear eval} &  \multirow{2}{*}{Supervised} \\
 &  &  &  &  ~~1\%~ &   ~~10\% &   ~100\% &  &  \\
\midrule
\multirow{4}{*}{50} & \multirow{2}{*}{1$\times$} & False &   \textbf{24} &        \textbf{57.9} &         \textbf{68.4} &          \textbf{76.3} &            \textbf{71.7} &         \textbf{76.6} \\
    &    & True  &   35 &        64.5 &         72.1 &          78.7 &            74.6 &         78.5 \\
    & \multirow{2}{*}{2$\times$} & False &   94 &        66.3 &         73.9 &          79.1 &            75.6 &         77.8 \\
    &    & True  &  140 &        70.6 &         77.0 &          81.3 &            77.7 &         79.3 \\
\multirow{4}{*}{101} & \multirow{2}{*}{1$\times$} & False &   43 &        62.1 &         71.4 &          78.2 &            73.6 &         78.0 \\
    &    & True  &   65 &        68.3 &         75.1 &          80.6 &            76.3 &         79.6 \\
    & \multirow{2}{*}{2$\times$} & False &  170 &        69.1 &         75.8 &          80.7 &            77.0 &         78.9 \\
    &    & True  &  257 &        73.2 &         78.8 &          82.4 &            79.0 &         80.1 \\
\multirow{4}{*}{152} & \multirow{2}{*}{1$\times$} & False &   58 &        64.0 &         73.0 &          79.3 &            74.5 &         78.3 \\
    &    & True  &   89 &        70.0 &         76.5 &          81.3 &            77.2 &         79.9 \\
    & \multirow{2}{*}{2$\times$} & False &  233 &        70.2 &         76.6 &          81.1 &            77.4 &         79.1 \\
    &    & True  &  354 &        74.2 &         79.4 &          82.9 &            79.4 &         80.4 \\
152 & 3$\times$ & True & \textbf{795} & {\textbf{74.9}} & {\textbf{80.1}} & {\textbf{83.1}} & {\textbf{79.8}} & {\textbf{80.5}} \\
\bottomrule
\end{tabular}
\end{table}

In order to study the effectiveness of big models, we train ResNet models by varying width and depth as well as whether or not to use selective kernels (SK)~\cite{li2019selective}.\footnote{Although we do not use grouped convolution in this work, we believe it can further improve parameter efficiency.} Whenever SK is used, we also use the ResNet-D~\cite{he2019bag} variant of ResNet. The smallest model is the standard ResNet-50, and biggest model is ResNet-152 ($3\times$+SK). 

Table~\ref{tab:big_ft_top1} compares self-supervised learning and supervised learning under different model sizes and evaluation protocols, including both fine-tuning and linear evaluation. We can see that increasing width and depth, as well as using SK, all improve the performance. These architectural manipulations have relatively limited effects for standard supervised learning (4\% differences in smallest and largest models), but for self-supervised models, accuracy can differ by as much as 8\% for linear evaluation, and 17\% for fine-tuning on 1\% of labeled images. We also note that ResNet-152 (3$\times$+SK) is only marginally better than ResNet-152 (2$\times$+SK), though the parameter size is almost doubled, suggesting that the benefits of width may have plateaued.

\begin{figure}[!t]
    \centering
    \small
    \begin{subfigure}{.33\textwidth}
      \centering
      \includegraphics[width=0.98\linewidth]{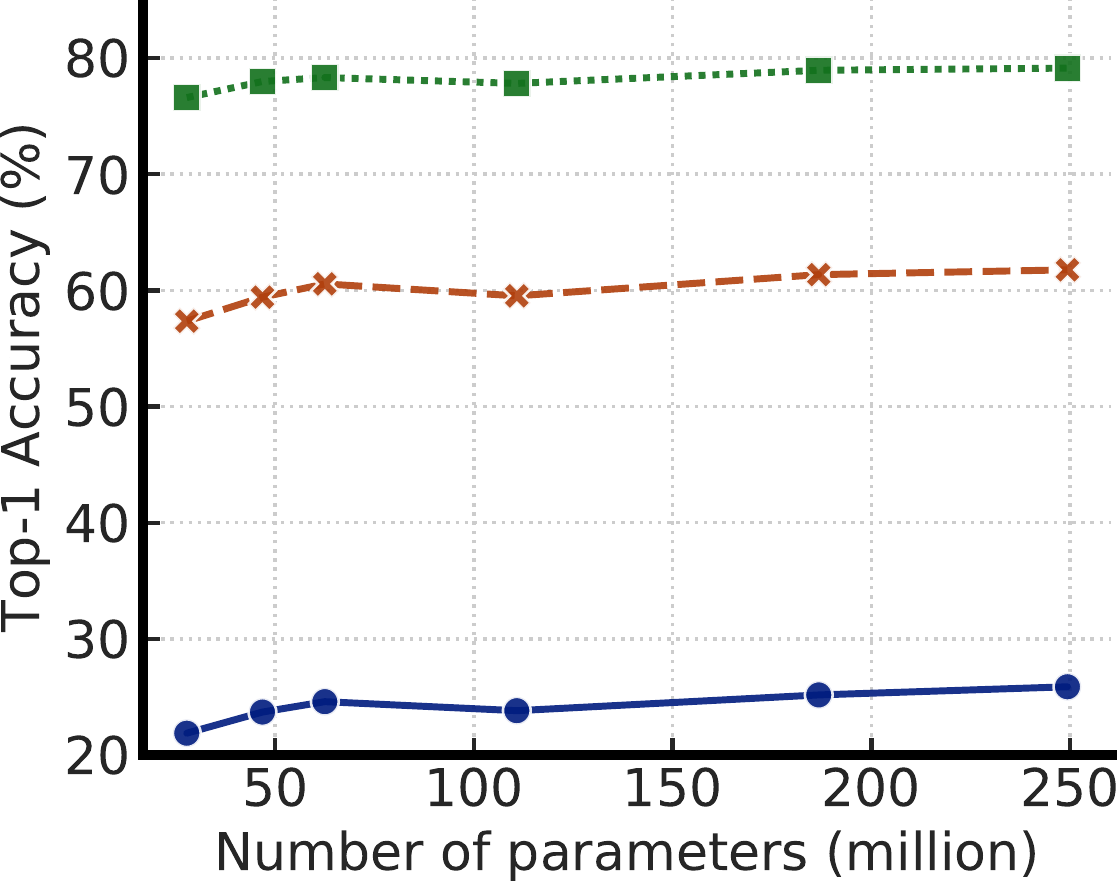}
      \caption{Supervised}
    \end{subfigure}\begin{subfigure}{.33\textwidth}
      \centering
      \includegraphics[width=0.98\linewidth]{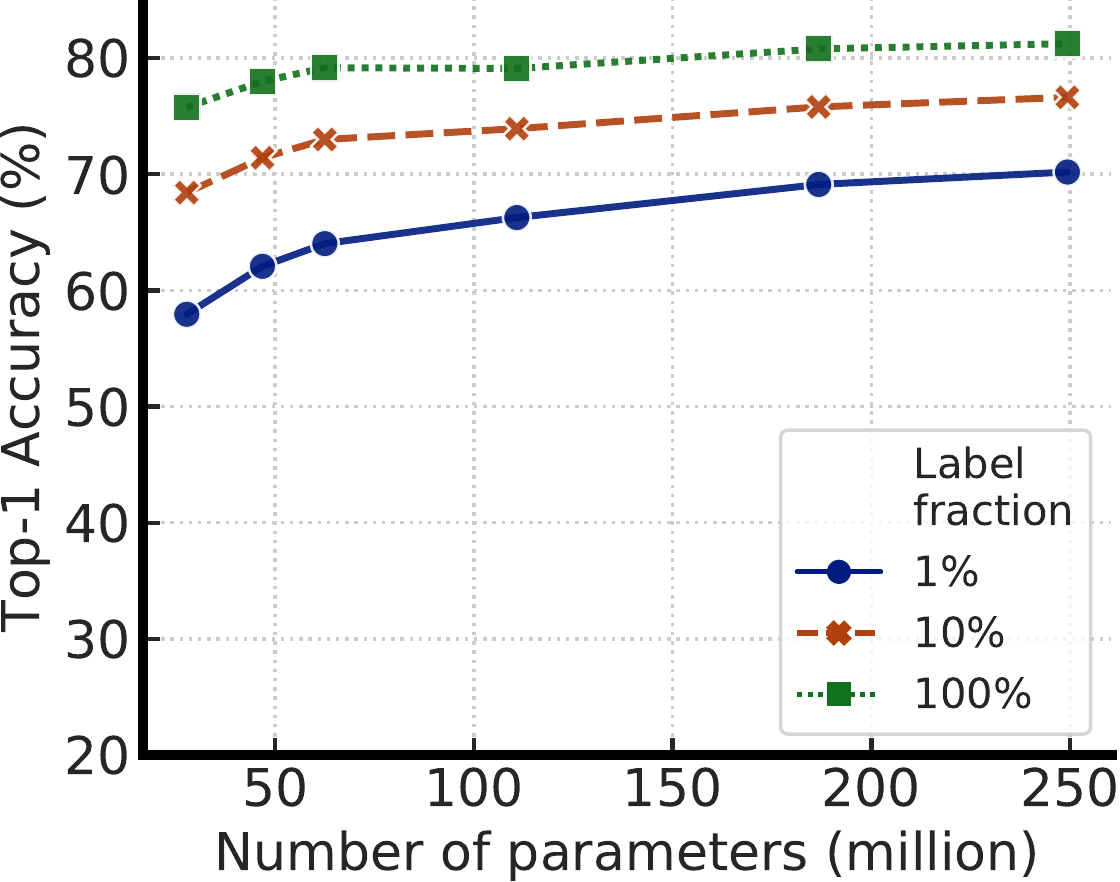}
      \caption{Semi-supervised}
    \end{subfigure}
    \begin{subfigure}{.33\textwidth}
      \centering
      \includegraphics[width=0.98\linewidth]{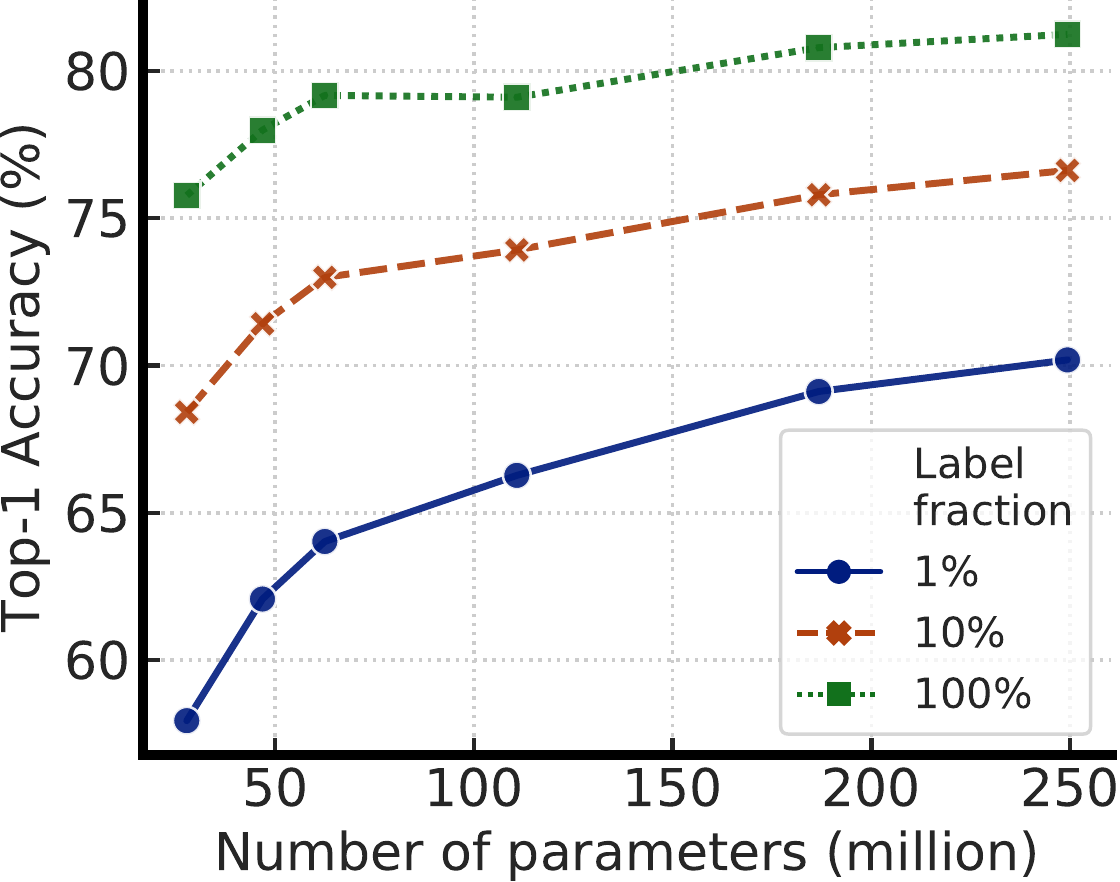}
      \caption{Semi-supervised (y-axis zoomed)}
    \end{subfigure}
    \caption{\label{fig:param_acc_simplified} Top-1 accuracy for supervised vs semi-supervised (SimCLRv2 fine-tuned) models of varied sizes on different label fractions. ResNets with depths of 50, 101, 152, width multiplier of $1\times$, $2\times$ (w/o SK) are presented here. For supervised models on 1\%/10\% labels, AutoAugment~\cite{cubuk2019autoaugment} and label smoothing~\cite{szegedy2016rethinking} are used. Increasing the size of SimCLRv2 models by  10$\times$, from ResNet-50 to ResNet-152 (2$\times$), improves label efficiency by 10$\times$.}
\end{figure}

Figure~\ref{fig:param_acc_simplified} shows the performance as model size and label fraction vary. These results show that bigger models are more label-efficient for \textit{both} supervised and semi-supervised learning, but gains appear to be larger for semi-supervised learning (more discussions in Appendix~\ref{app:param_acc}). Furthermore, it is worth pointing out that although bigger models are better, some models (e.g. with SK) are more parameter efficient than others (Appendix~\ref{app:param_efficiency}), suggesting that searching for better architectures is helpful.

\subsection{Bigger/Deeper Projection Heads Improve Representation Learning}

\begin{figure}[h]
    \centering
    \small
    \begin{subfigure}{.75\textwidth}
      \centering
      \includegraphics[width=\linewidth]{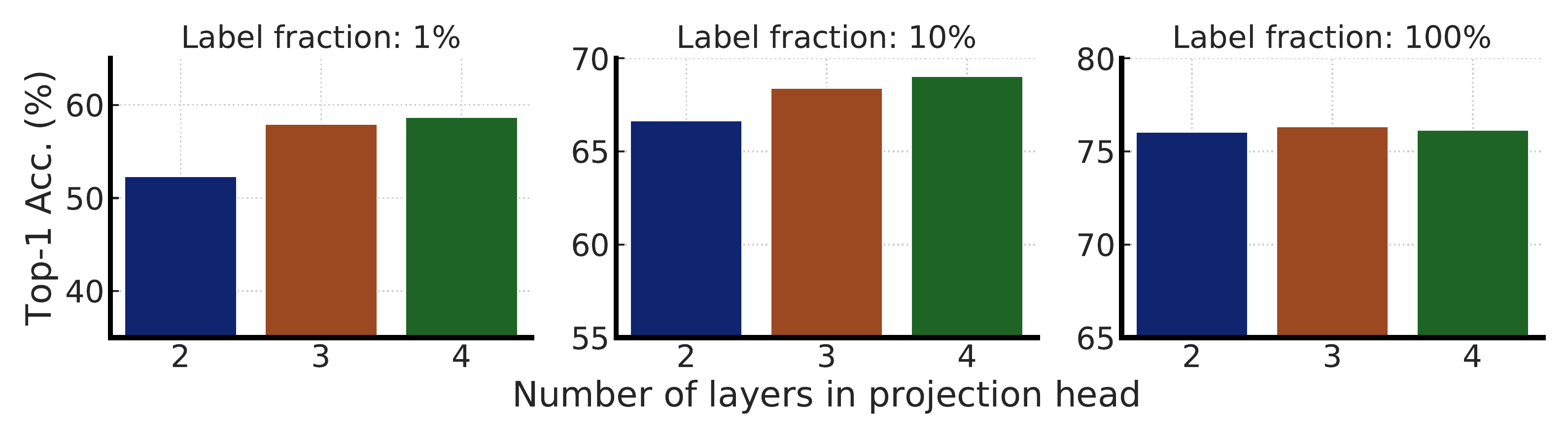}
      \caption{\label{fig:proj_head_study_r50_a} Effect of projection head's depth when fine-tuning from optimal middle layer.}
    \end{subfigure}
    \\
    \begin{subfigure}{.75\textwidth}
      \centering
      \includegraphics[width=\linewidth]{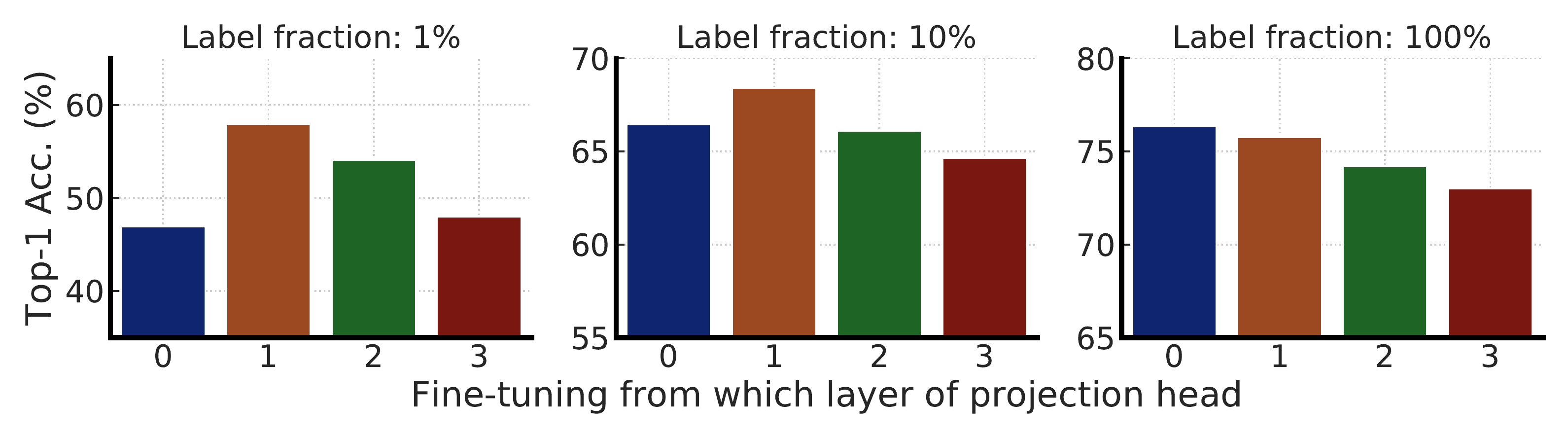}     \caption{\label{fig:proj_head_study_r50_b}Effect of fine-tuning from middle of a 3-layer projection head (0 is SimCLR).}
    \end{subfigure}
    \caption{\label{fig:proj_head_study_r50} Top-1 accuracy via fine-tuning under different projection head settings and label fractions (using ResNet-50).}
\end{figure}

To study the effects of projection head for fine-tuning, we pretrain ResNet-50 using SimCLRv2 with different numbers of projection head layers (from 2 to 4 fully connected layers), and examine performance when fine-tuning from different layers of the projection head. We find that using a deeper projection head during pretraining is better when fine-tuning from the optimal layer of projection head (Figure~\ref{fig:proj_head_study_r50_a}), and this optimal layer is typically the first layer of projection head rather than the input (0\textsuperscript{th} layer), especially when fine-tuning on fewer labeled examples (Figure~\ref{fig:proj_head_study_r50_b}).

It is also worth noting that when using bigger ResNets, the improvements from having a deeper projection head are smaller (see Appendix~\ref{app:proj_head}). In our experiments, wider ResNets also have wider projection heads, since the width multiplier is applied to both. Thus, it is possible that increasing the depth of the projection head has limited effect when the projection head is already relatively wide.

When varying architecture, the accuracy of fine-tuned models is correlated with the accuracy of linear evaluation (see Appendix~\ref{app:linear_ft_corr}). Although we use the input of the projection head for linear classification, we find that the correlation is higher when fine-tuning from the optimal middle layer of the projection head than when fine-tuning from the projection head input.

\subsection{Distillation Using Unlabeled Data Improves Semi-Supervised Learning}

Distillation typically involves both a distillation loss that encourages the student to match a teacher and an ordinary supervised cross-entropy loss on the labels (Eq.~\ref{eq:distill_plus}). In Table~\ref{tab:dist_abl}, we demonstrate the importance of using unlabeled examples when training with the distillation loss. Furthermore, using the distillation loss alone (Eq.~\ref{eq:distill}) works almost as well as balancing distillation and label losses (Eq.~\ref{eq:distill_plus}) when the labeled fraction is small. For simplicity, Eq.~\ref{eq:distill} is our default for all other experiments.

\begin{table}[!t]
    \small
    \centering
    \caption{\label{tab:dist_abl}Top-1 accuracy of a ResNet-50 trained on different types of targets. For distillation, the temperature is set to 1.0, and the teacher is ResNet-50 (2$\times$+SK), which gets 70.6\% with 1\% of the labels and 77.0\% with 10\%, as shown in in Table~\ref{tab:big_ft_top1}. The distillation loss (Eq.~\ref{eq:distill}) does not use label information. Neither strong augmentation nor extra regularization are used.}
    \begin{tabular}{lcc}
    \toprule
    \multirow{2}{*}{Method} & \multicolumn{2}{c}{Label fraction}\\
     & 1\% & 10\% \\ \midrule
    Label only &  12.3 & 52.0 \\
    Label + distillation loss (on labeled set) & 23.6 & 66.2 \\
Label + distillation loss (on labeled+unlabeled sets) &  69.0 & 75.1 \\  Distillation loss (on labeled+unlabeled sets; our default) & 68.9 & 74.3 \\  \bottomrule
    \end{tabular}
\end{table}

\begin{figure}[h]
    \centering
    \small
    \begin{subfigure}{.42\textwidth}
      \centering
      \includegraphics[width=0.98\linewidth]{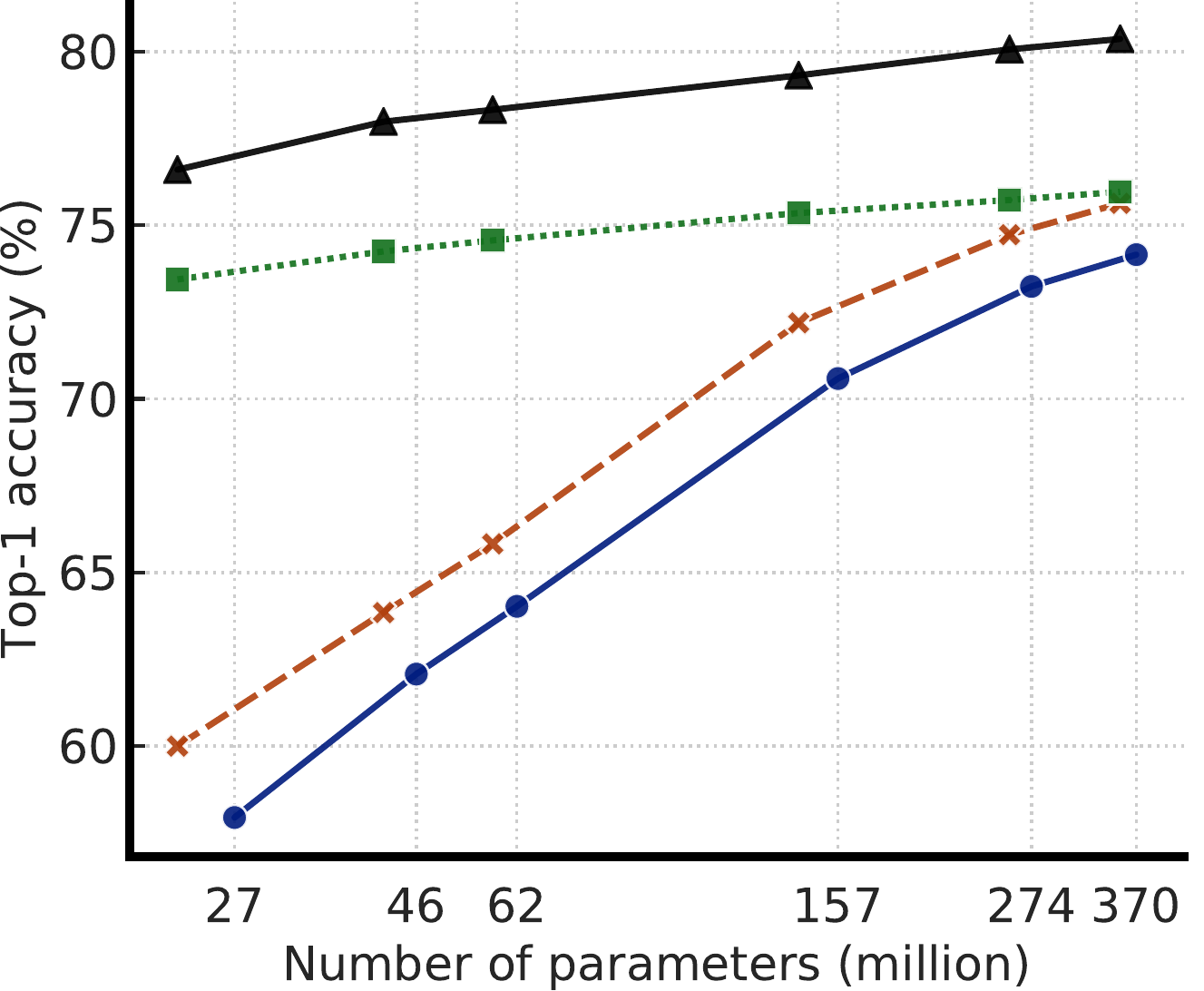}
      \caption{Label fraction 1\%}
    \end{subfigure}
    ~~~~~~~~~~
    \begin{subfigure}{.42\textwidth}
      \centering
      \includegraphics[width=0.98\linewidth]{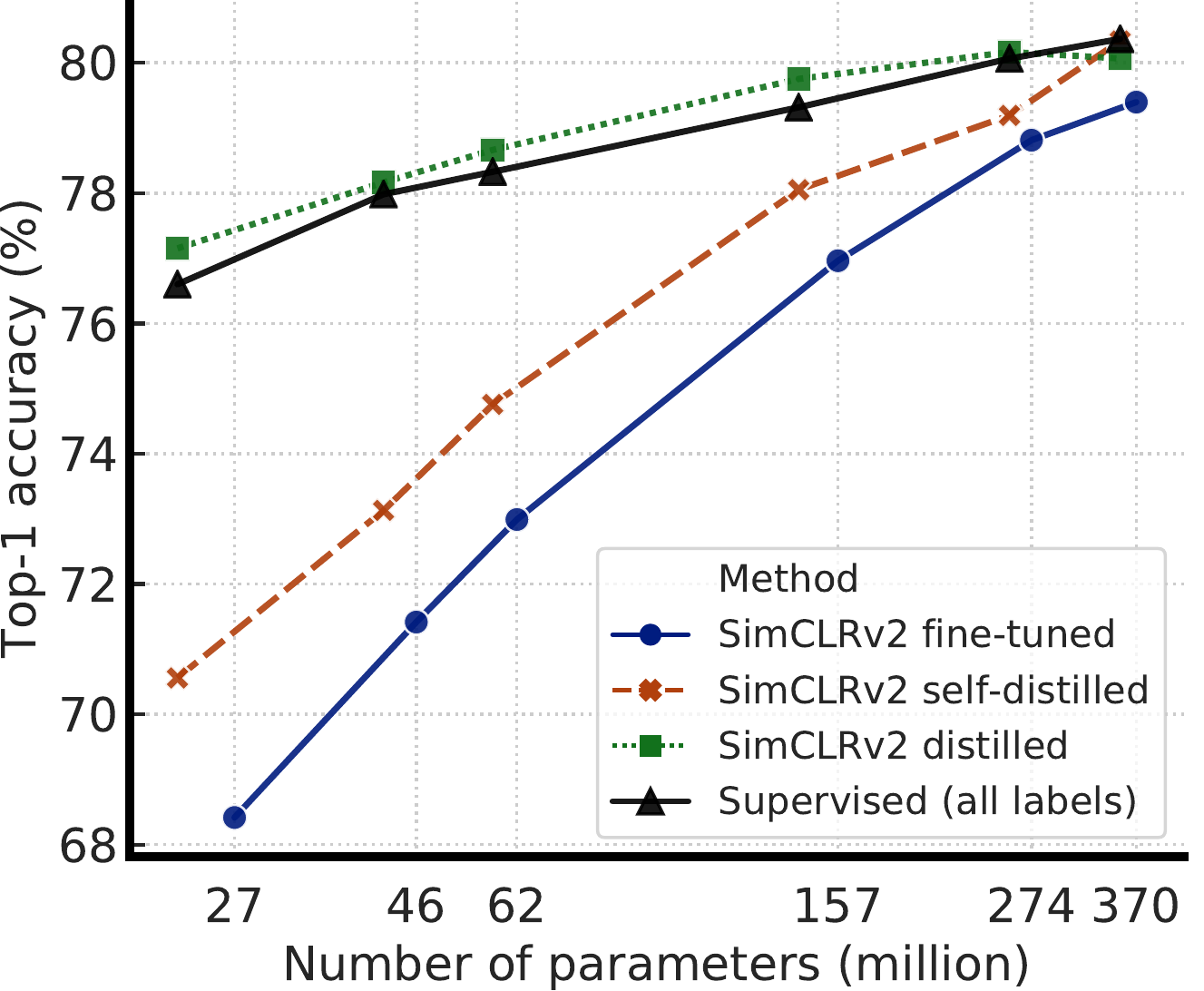}
      \caption{Label fraction 10\%}
    \end{subfigure}
    \caption{\label{fig:dist_curve} Top-1 accuracy of distilled SimCLRv2 models compared to the fine-tuned models as well as supervised learning with all labels. The self-distilled student has the same ResNet as the teacher (without MLP projection head). The distilled student is trained using the self-distilled ResNet-152 (2$\times$+SK) model, which is the largest model included in this figure.}
\end{figure}

Distillation with unlabeled examples improves fine-tuned models in two ways, as shown in Figure~\ref{fig:dist_curve}: (1) when the student model has a smaller architecture than the teacher model, it improves the model efficiency by transferring task-specific knowledge to a student model, (2) even when the student model has the same architecture as the teacher model (excluding the projection head after ResNet encoder), self-distillation can still meaningfully improve the semi-supervised learning performance. To obtain the best performance for smaller ResNets, the big model is self-distilled before distilling it to smaller models.

\begin{table}[!t]
\small
\centering
\caption{\label{tab:sota_more} ImageNet accuracy of models trained under semi-supervised settings. For our methods, we report results with distillation after fine-tuning. For our smaller models, we use self-distilled ResNet-152 ($3\times$+SK) as the teacher.}
\begin{tabular}{llcccc}
\toprule
\multirow{3}{*}{Method} & \multirow{3}{*}{Architecture} & \multicolumn{2}{c}{Top-1} & \multicolumn{2}{c}{Top-5} \\
& & \multicolumn{2}{c}{Label fraction} &  \multicolumn{2}{c}{Label fraction} \\
& & 1\% & 10\% & 1\% & 10\% \\
\midrule
Supervised baseline~\cite{zhai2019s4l}      & ResNet-50  &  25.4 & 56.4 & 48.4 & 80.4 \\ \midrule
\multicolumn{4}{l}{\textit{Methods using unlabeled data in a task-specific way:}}         \\ 
Pseudo-label~\cite{lee2013pseudo,zhai2019s4l}          & ResNet-50 &  -& -& 51.6 & 82.4\\
VAT+Entropy Min.~\cite{grandvalet2005semi,miyato2018virtual,zhai2019s4l}          & ResNet-50 &  -& -& 47.0 & 83.4 \\
Mean teacher~\cite{tarvainen2017mean} &  ResNeXt-152 & - & - & - & 90.9 \\
UDA (w. RandAug)~\cite{xie2019unsupervised}                        & ResNet-50 &  - & 68.8 & - & 88.5\\
FixMatch (w. RandAug)~\cite{sohn2020fixmatch}         & ResNet-50 & - & 71.5 & - & 89.1\\
S4L (Rot+VAT+Entropy Min.)~\cite{zhai2019s4l}          & ResNet-50 (4$\times$) &  - &  73.2 & - & 91.2\\
MPL (w. RandAug)~\cite{pham2020meta} & ResNet-50 & - & 73.8 & - & - \\
CowMix~\cite{french2020milking} & ResNet-152 & - & 73.9 & - & 91.2 \\
\midrule
\multicolumn{4}{l}{\textit{Methods using unlabeled data in a task-agnostic way:}}         \\ 
InstDisc~\cite{wu2018unsupervised}  & ResNet-50 & -& - &39.2 & 77.4\\
BigBiGAN~\cite{donahue2019large} & RevNet-50 ($4\times$)             & - &- & 55.2  & 78.8  \\
PIRL~\cite{misra2019self} & ResNet-50             & -& -& 57.2  & 83.8  \\
CPC v2~\cite{henaff2019data} & ResNet-161($*$)            & 52.7  &  73.1 & 77.9 & 91.2\\
SimCLR~\cite{chen2020simple}   & ResNet-50        & 48.3 & 	65.6  & 75.5 & 87.8 \\
SimCLR~\cite{chen2020simple}   & ResNet-50 ($2\times$)           & 58.5  &  71.7  &  83.0 & 91.2\\
SimCLR~\cite{chen2020simple}    & ResNet-50 ($4\times$)           & {63.0} &  {74.4} & 85.8 & 92.6 \\
BYOL~\cite{grill2020bootstrap} (concurrent work)    & ResNet-50 & 53.2 & 68.8 & 78.4 & 89.0 \\
BYOL~\cite{grill2020bootstrap}  (concurrent work)   & ResNet-200 ($2\times$)           & 71.2 & 77.7 & 89.5 & 93.7 \\
\midrule
\multicolumn{4}{l}{\textit{Methods using unlabeled data in both ways:}}         \\
SimCLRv2 distilled (ours)    & ResNet-50         & {73.9} &  {77.5} &  91.5 & 93.4\\
SimCLRv2 distilled (ours)    & ResNet-50 ($2\times$+SK)         & {75.9}   &  {80.2} &  93.0 & 95.0 \\
SimCLRv2 self-distilled (ours)    & ResNet-152 ($3\times$+SK)         & \textbf{76.6}   &  \textbf{80.9} & \textbf{93.4} & \textbf{95.5}\\
\bottomrule
\end{tabular}
\end{table}

We compare our best models with previous state-of-the-art semi-supervised learning methods (and a concurrent work~\cite{grill2020bootstrap}) on ImageNet in Table~\ref{tab:sota_more}. Our approach greatly improves upon previous results, for both small and big ResNet variants.

 \section{Related work}

\textbf{Task-agnostic use of unlabeled data.} Unsupervised or self-supervised pretraining followed by supervised fine-tuning on a few labeled examples has been extensively used in natural language processing~\cite{kiros2015skip,dai2015semi,radford2018improving,peters2018deep,devlin2018bert}, but has only shown promising results in computer vision very recently~\cite{henaff2019data,he2019momentum,chen2020simple}. 
Our work builds upon recent success on contrastive learning of visual representations~\cite{dosovitskiy2014discriminative,oord2018representation,wu2018unsupervised,hjelm2018learning,bachman2019learning,henaff2019data,tian2019contrastive,misra2019self,he2019momentum,li2020prototypical,tian2020makes,chen2020simple}, a sub-area within self-supervised learning. These contrastive learning based approaches learn representations in a discriminative fashion instead of a generative one as in~\cite{hinton2006fast,kingma2013auto,goodfellow2014generative,donahue2019large,chen2019self}. There are other approaches to self-supervised learning that are based on handcrafted pretext tasks~\cite{doersch2015unsupervised,zhang2016colorful,noroozi2016unsupervised,gidaris2018unsupervised,kolesnikov2019revisiting,goyal2019scaling}. We also note a concurrent work on advancing self-supervised pretraining without using negative examples~\cite{grill2020bootstrap}, which we also compare against in Table~\ref{tab:sota_more}.
Our work also extends the ``unsupervised pretrain, supervised fine-tune'' paradigm by combining it with (self-)distillation~\cite{bucilua2006model,hinton2015distilling,lee2013pseudo} using unlabeled data.

\textbf{Task-specific use of unlabeled data.} Aside from the representation learning paradigm, there is a large and diverse set of approaches for semi-supervised learning, we refer readers to~\cite{chapelle2006semi,zhu2009introduction,oliver2018realistic} for surveys of classical approaches. Here we only review methods closely related to ours (especially within computer vision). One family of highly relevant methods are based on pseudo-labeling~\cite{lee2013pseudo,sohn2020fixmatch} or self-training~\cite{xie2019self,yalniz2019billion,zou2019confidence}. The main differences between these methods and ours are that our initial / teacher model is trained using SimCLRv2 (with unsupervised pretraining and supervised fine-tuning), and the student models can also be smaller than the initial / teacher model. Furthermore, we use temperature scaling instead of confidence-based thresholding, and we do not use strong augmentation for training the student. Another family of methods are based on label consistency regularization~\cite{bachman2014learning,laine2016temporal,sajjadi2016regularization,tarvainen2017mean,xie2019unsupervised,berthelot2019mixmatch,verma2019interpolation,sohn2020fixmatch}, where unlabeled examples are directly used as a regularizer to encourage task prediction consistency. Although in SimCLRv2 pretraining, we maximize the agreement/consistency of representations of the same image under different augmented views, there is \textit{no} supervised label utilized to compute the loss, a crucial difference from label consistency losses.

 \section{Discussion}
\label{sec:discussion}

In this work, we present a simple framework for semi-supervised ImageNet classification in three steps: unsupervised pretraining, supervised fine-tuning, and distillation with unlabeled data. Although similar approaches are common in NLP, we demonstrate that this approach can also be a surprisingly strong baseline for semi-supervised learning in computer vision, outperforming the state-of-the-art by a large margin.

We observe that bigger models can produce larger improvements with fewer labeled examples. We primarily study this phenomenon on ImageNet, but we observe similar results on CIFAR-10, a much smaller dataset (see appendix~\ref{app:cifar10}). The effectiveness of big models have been demonstrated on supervised learning~\cite{sun2017revisiting,hestness2017deep,kaplan2020scaling,mahajan2018exploring}, fine-tuning supervised models on a few examples~\cite{kolesnikov2019large}, and unsupervised learning on language~\cite{devlin2018bert,raffel2019exploring,radford2019language,brown2020language}.
However, it is still somewhat surprising that bigger models, which could easily overfit with few labeled examples, can generalize much better.
With task-agnostic use of unlabeled data, we conjecture bigger models can learn more general features, which increases the chances of learning task-relevant features. However, further work is needed to gain a better understanding of this phenomenon. Beyond model size, we also see the importance of increasing parameter efficiency as the other important dimension of improvement.

Although big models are important for pretraining and fine-tuning, given a specific task, such as classifying images into 1000 ImageNet classes, we demonstrate that task-agnostically learned general representations can be distilled into a more specialized and compact network using unlabeled examples. We simply use the teacher to impute labels for the unlabeled examples for this purpose, without using noise, augmentation, confidence thresholding, or consistency regularization. When the student network has the same or similar architecture as the teacher, this process can consistently improve the classification performance. We believe our framework can benefit from better approaches to leverage the unlabeled data for improving and transferring task-specific knowledge. We also recognize that ImageNet is a well-curated dataset, and may not reflect all real-world applications of semi-supervised learning. Thus, a potential future direction is to explore wider range of real datasets.

 \section{Broader Impact}

The findings described in this paper can potentially be harnessed to improve accuracy in any application of computer vision where it is more expensive or difficult to label additional data than to train larger models. Some such applications are clearly beneficial to society. For example, in medical applications where acquiring high-quality labels requires careful annotation by clinicians, better semi-supervised learning approaches can potentially help save lives. Applications of computer vision to agriculture can increase crop yields, which may help to improve the availability of food. However, we also recognize that our approach could become a component of harmful surveillance systems. Moreover, there is an entire industry built around human labeling services, and technology that reduces the need for these services could lead to a short-term loss of income for some of those currently employed or contracted to provide labels.

\section*{Acknowledgements}
We would like to thank David Berthelot, Han Zhang, Lala Li, Xiaohua Zhai, Lucas Beyer, Alexander Kolesnikov for their helpful feedback on the draft. We are also grateful for general support from Google Research teams in Toronto and elsewhere.

{\small
\bibliography{content/ref}
\bibliographystyle{unsrtnat}}
\clearpage\newpage
\counterwithin{figure}{section}
\counterwithin{table}{section}
\appendix

\section{When Do Bigger Models Help More?}
\label{app:param_acc}

\begin{figure}[h]
    \centering
    \small
    \begin{subfigure}{.32\textwidth}
      \centering
      \caption{\label{semi_param_more_1}Supervised}
      \includegraphics[width=0.98\linewidth]{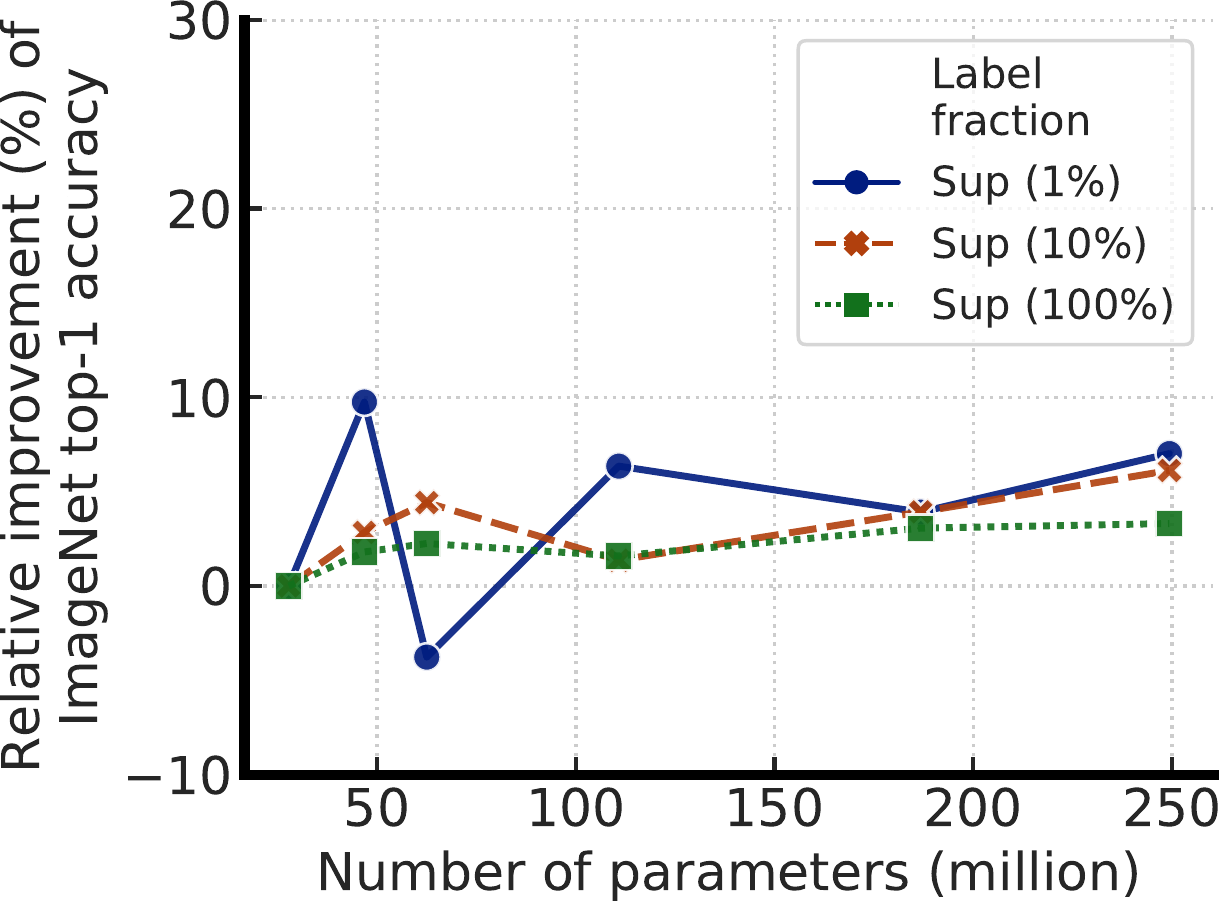}
    \end{subfigure}
    \begin{subfigure}{.32\textwidth}
      \centering
      \caption{\label{semi_param_more_2}Supervised (auto-augment+LS)}
      \includegraphics[width=0.98\linewidth]{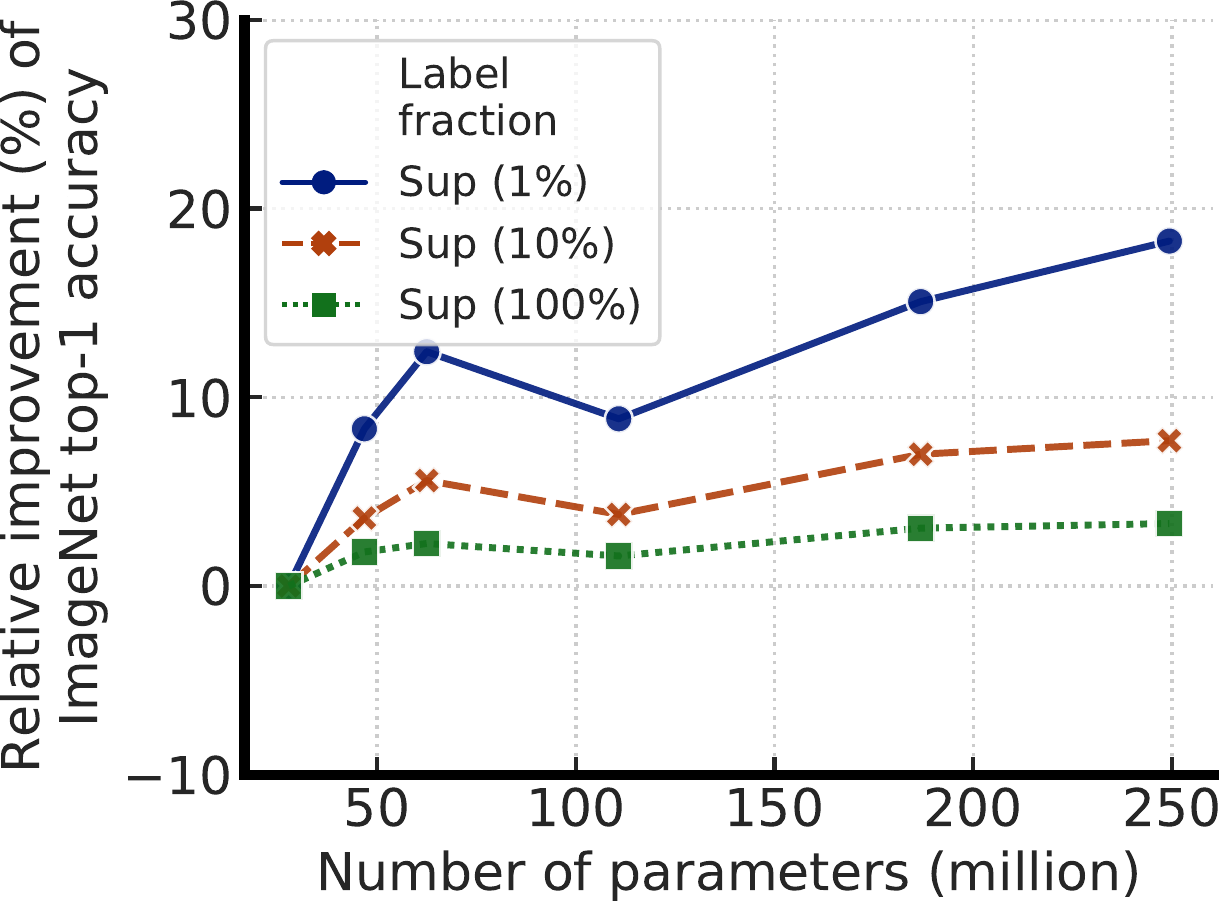}
    \end{subfigure}
    \begin{subfigure}{.32\textwidth}
      \centering
      \caption{\label{semi_param_more_3}Semi-supervised}
      \includegraphics[width=0.98\linewidth]{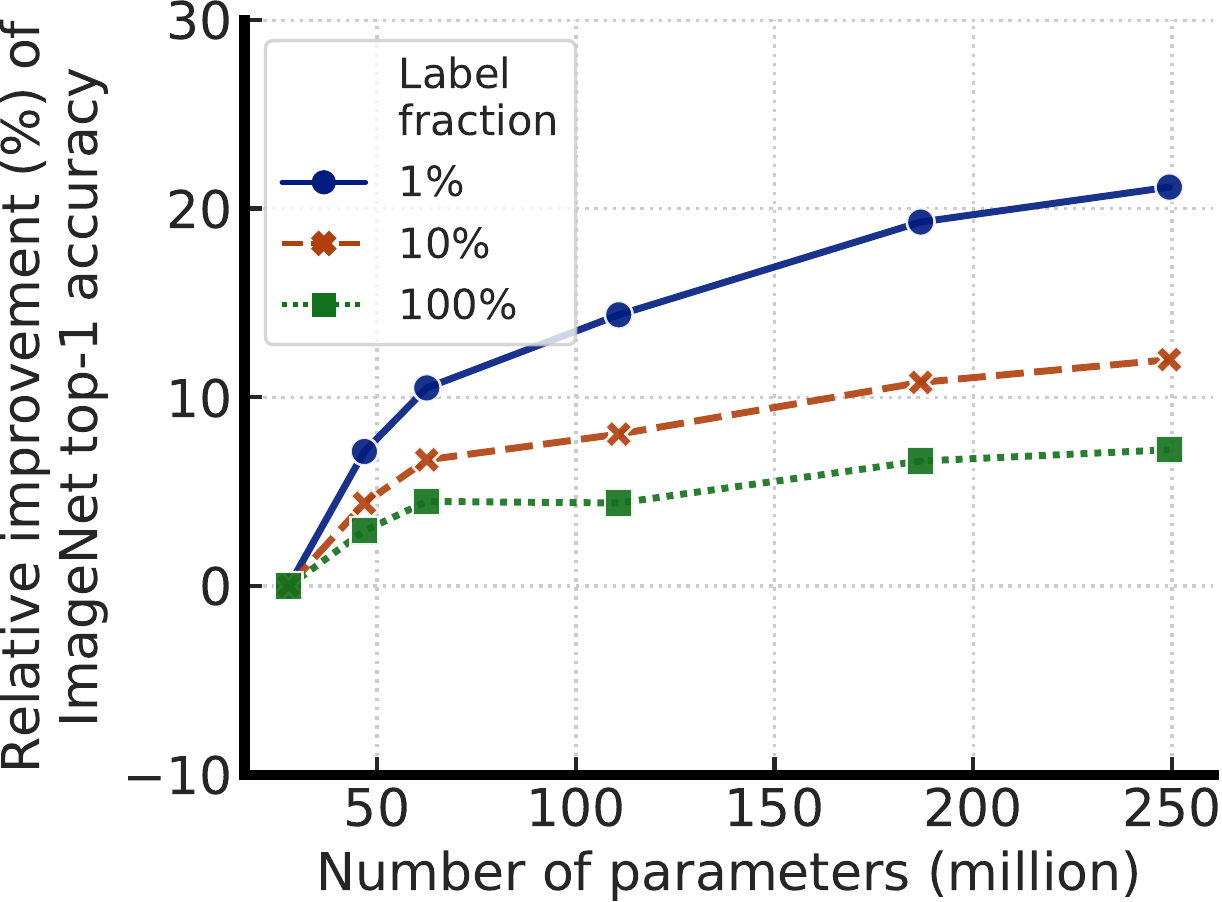}
    \end{subfigure}\caption{\label{fig:semi_param_more} Relative improvement (top-1) when model size is increased. (a) supervised learning without extra regularization, (b), Supervised learning with auto-augmentation~\cite{cubuk2019autoaugment} and label smoothing~\cite{szegedy2015going} are applied for 1\%/10\% label fractions, (c) semi-supervised learning by fine-tuning SimCLRv2.}
\end{figure}

Figure~\ref{fig:semi_param_more} shows relative improvement by increasing the model size under different amount of labeled examples. Both supervised learning and semi-supervised learning (i.e. SimCLRv2) seem to benefit from having bigger models. The benefits are larger when (1) regularization techniques (such as augmentation, label smoothing) are used, or (2) the model is pretrained using unlabeled examples. It is also worth noting that these results may reflect a ``ceiling effect'': as the performance gets closer to the ceiling, the improvement becomes smaller. 

\section{Parameter Efficiency Also Matters}
\label{app:param_efficiency}

\begin{figure}[h]
    \centering
    \small
    \begin{subfigure}{.4\textwidth}
      \centering
      \includegraphics[width=0.98\linewidth]{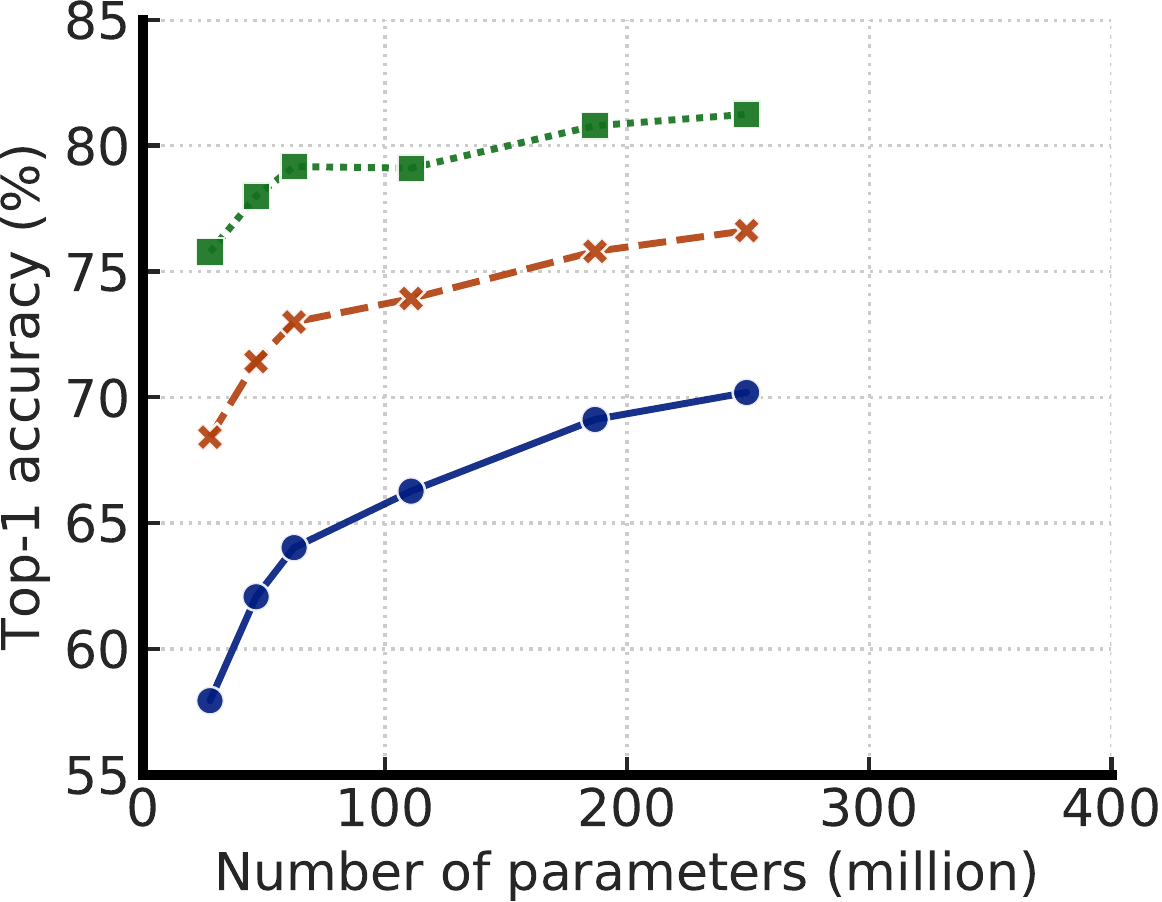}
      \caption{Models without SK~\cite{li2019selective}}
    \end{subfigure}\begin{subfigure}{.4\textwidth}
      \centering
      \includegraphics[width=0.98\linewidth]{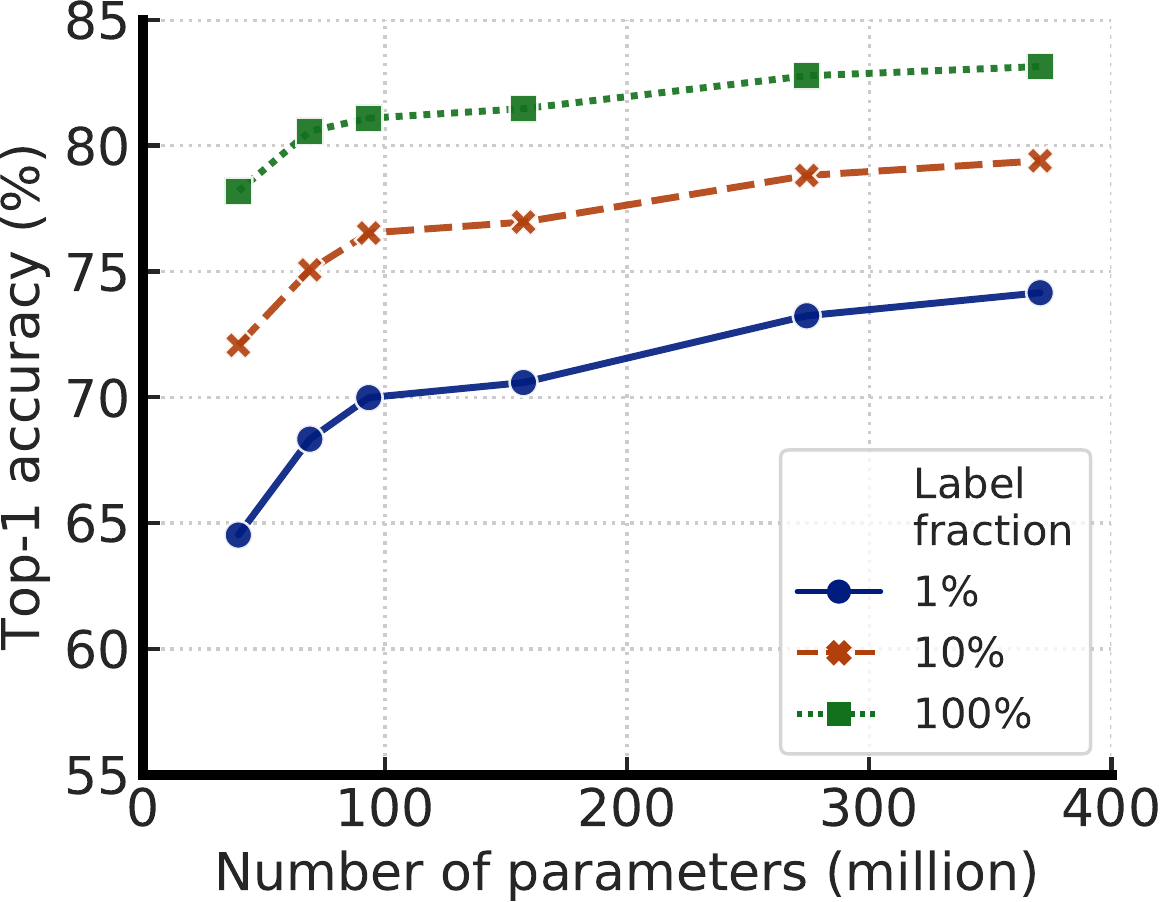}
      \caption{Models with SK~\cite{li2019selective}}
    \end{subfigure}
    \caption{\label{fig:param_efficiency} Top-1 accuracy of fined-tuned SimCLRv2 models of different sizes on three label fractions. ResNets with depth in $\{50,101,152\}$, width in $\{1\times, 2\times\}$ are included here. Parameter efficiency also plays an important role. For fine-tuning on 1\% of labels, SK is much more efficient.}
\end{figure}

Figure~\ref{fig:param_efficiency} shows the top-1 accuracy of fine-tuned SimCLRv2 models of different sizes. It shows that (1) bigger models are better, but (2) with SK~\cite{li2019selective}, better performance can be achieved with the same parameter count. It is worth to note that, in this work, we do not leverage group convolution for SK~\cite{li2019selective} and we use only $3\times 3$ kernels. We expect further improvement in terms of parameter efficiency if group convolution is utilized.

\section{The Correlation Between Linear Evaluation and Fine-tuning}
\label{app:linear_ft_corr}

\begin{figure}[h]
    \centering
    \includegraphics[width=0.8\linewidth]{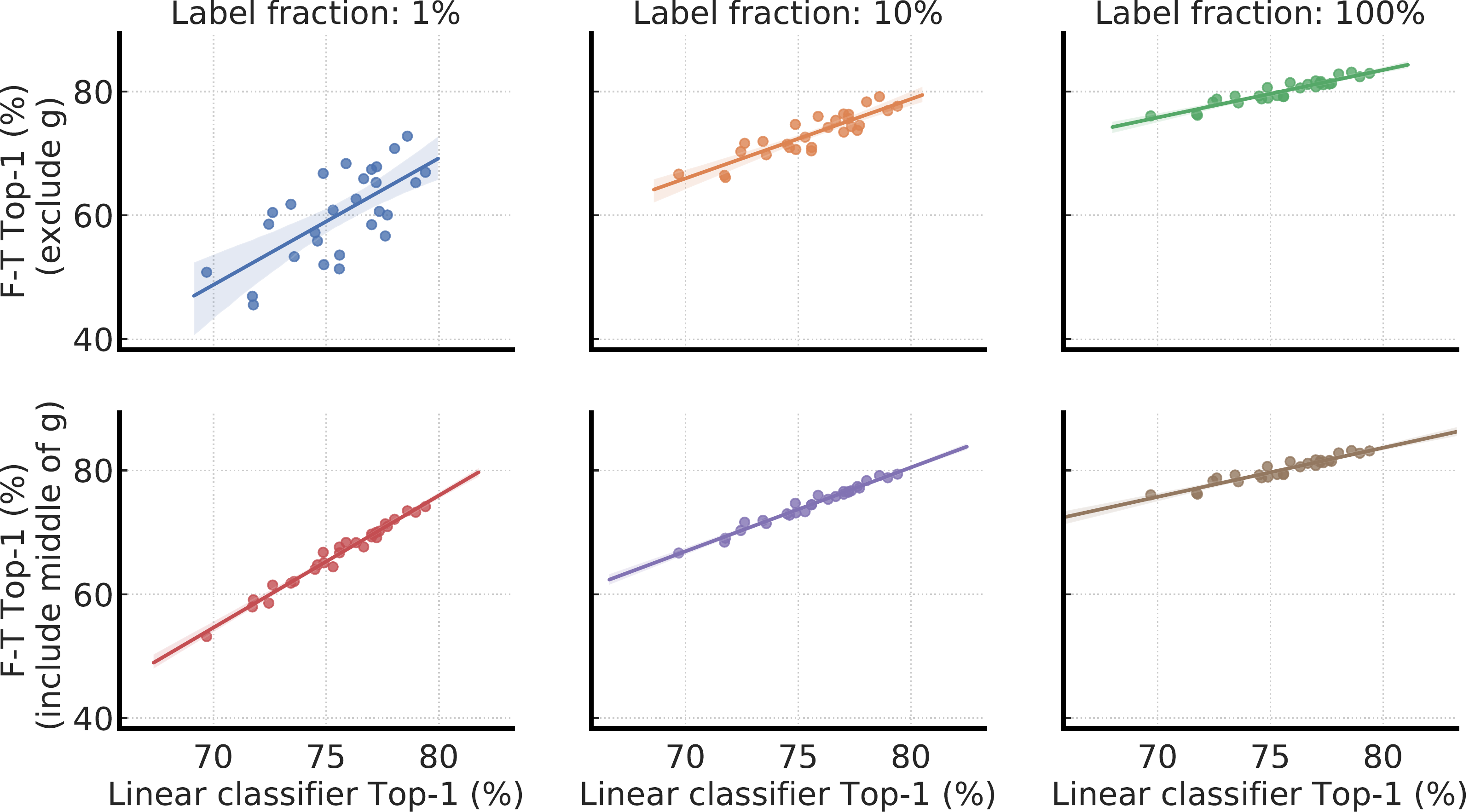}
    \caption{\label{fig:correlation_ft_linear} The effects of projection head for the correlation between fine-tuning and linear evaluation. When allowing fine-tuning from the middle of the projection head, the linear correlation becomes stronger. Furthermore, as label fraction increases, the slope is decreasing. The points here are from the variants of ResNets with depth in $\{50, 101, 152\}$, width in $\{1\times, 2\times\}$, and with/without SK.}
\end{figure}

Most existing work~\cite{wu2018unsupervised,henaff2019data,bachman2019learning,misra2019self,he2019momentum,chen2020simple} on self-supervised learning leverages linear evaluation as a main metric for evaluating representation quality, and it is not clear how it correlates with semi-supervised learning through fine-tuning. Here we further study the correlation of fine-tuning and linear evaluation (the linear classifier is trained on the ResNet output instead of some middle layer of projection head). Figure~\ref{fig:correlation_ft_linear} shows the correlation under two different fine-tuning strategies: fine-tuning from the input of projection head, or fine-tuning from a middle layer of projection head. We observe that overall there is a linear correlation. When fine-tuned from a middle layer of the projection head, we observe a even stronger linear correlation. Additionally, we notice the slope of correlation becomes smaller as number of labeled images for fine-tuning increases .

\section{The Impact of Memory}
\label{app:moco}

\begin{figure}[h]
    \centering
    \includegraphics[width=\linewidth]{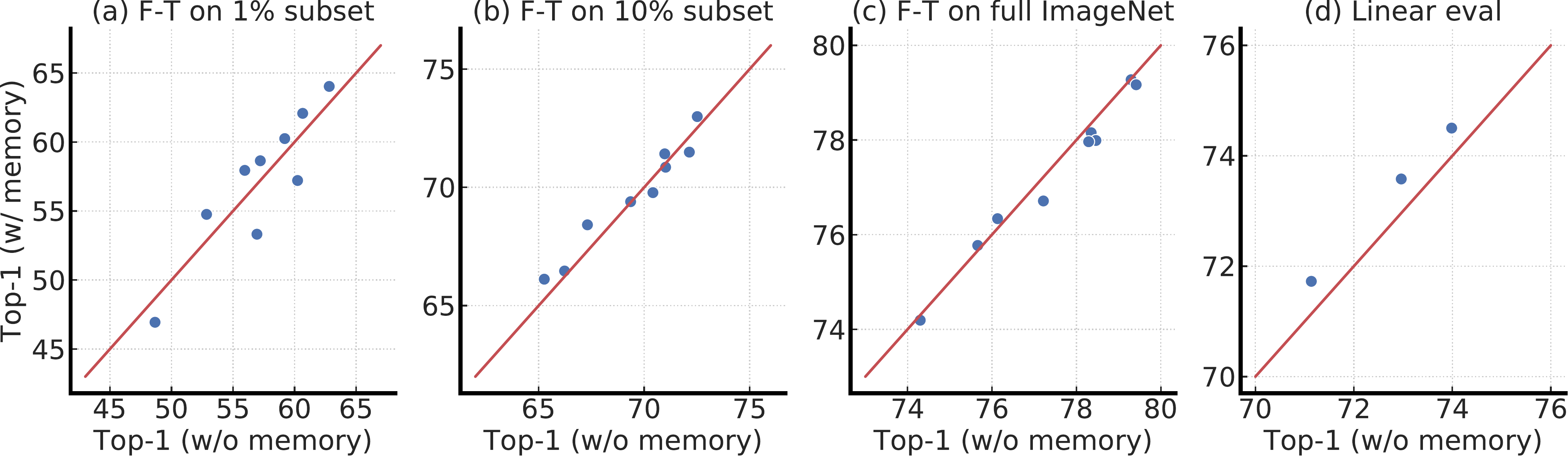}
    \caption{\label{fig:abt_memory} Top-1 results of ResNet-50, ResNet-101, and ResNet-152 trained with or without memory.}
\end{figure}

Figure~\ref{fig:abt_memory} shows the top-1 comparisons for SimCLRv2 models trained with or without memory (MoCo)~\cite{he2019momentum}. Memory provides modest advantages in terms of linear evaluation and fine-tuning with 1\% of the labels; the improvement is around 1\%. We believe the reason that memory only provides marginal improvement is that we already use a big batch size (i.e. 4096).

\section{The Impact of Projection Head Under Different Model Sizes}
\label{app:proj_head}

\begin{figure}[h]
    \centering
    \small
    \begin{subfigure}{.3\textwidth}
      \centering
      \includegraphics[width=0.98\linewidth]{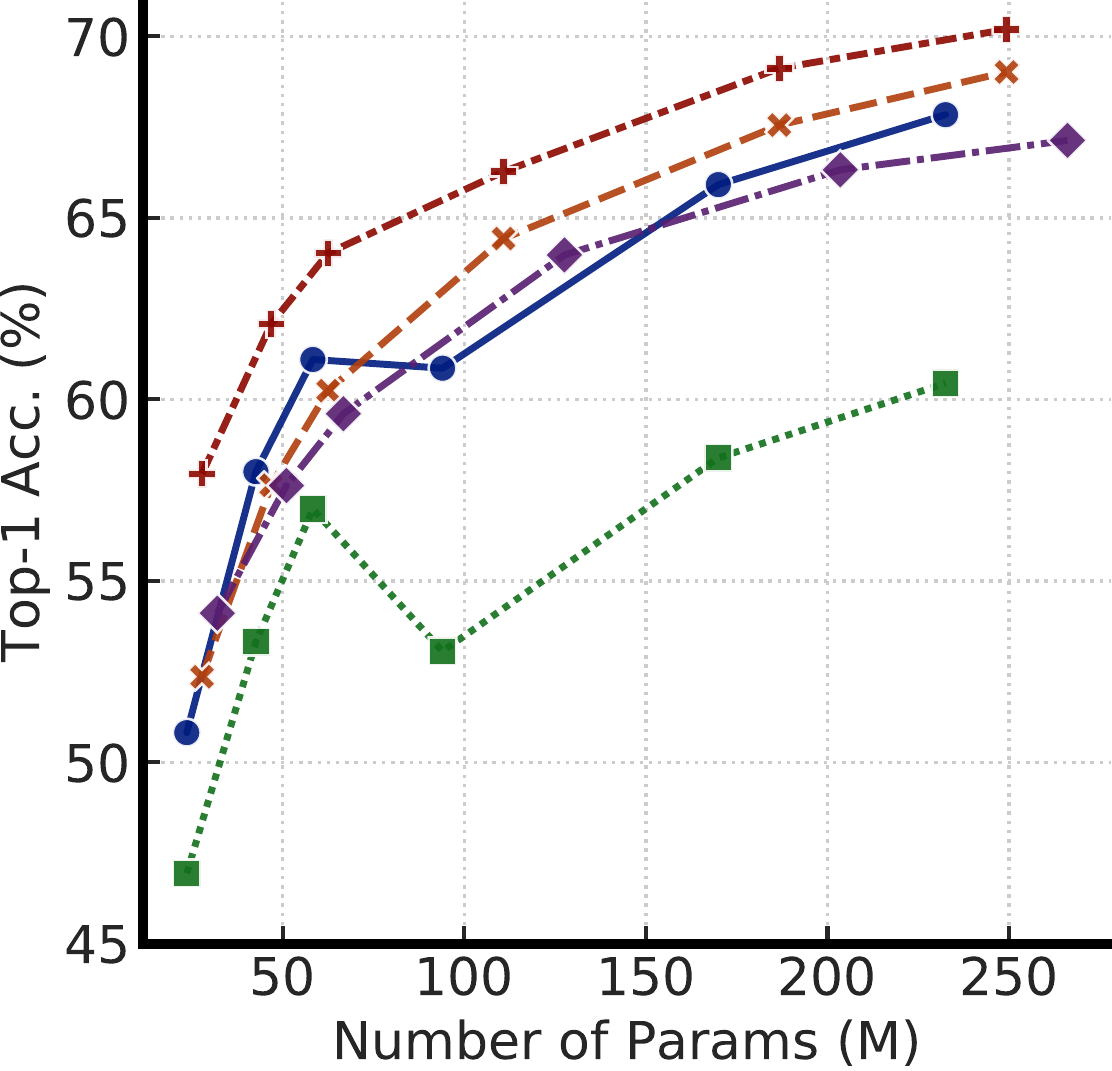}
      \caption{w/o SK (label fraction: 1\%)}
    \end{subfigure}\begin{subfigure}{.3\textwidth}
      \centering
      \includegraphics[width=0.98\linewidth]{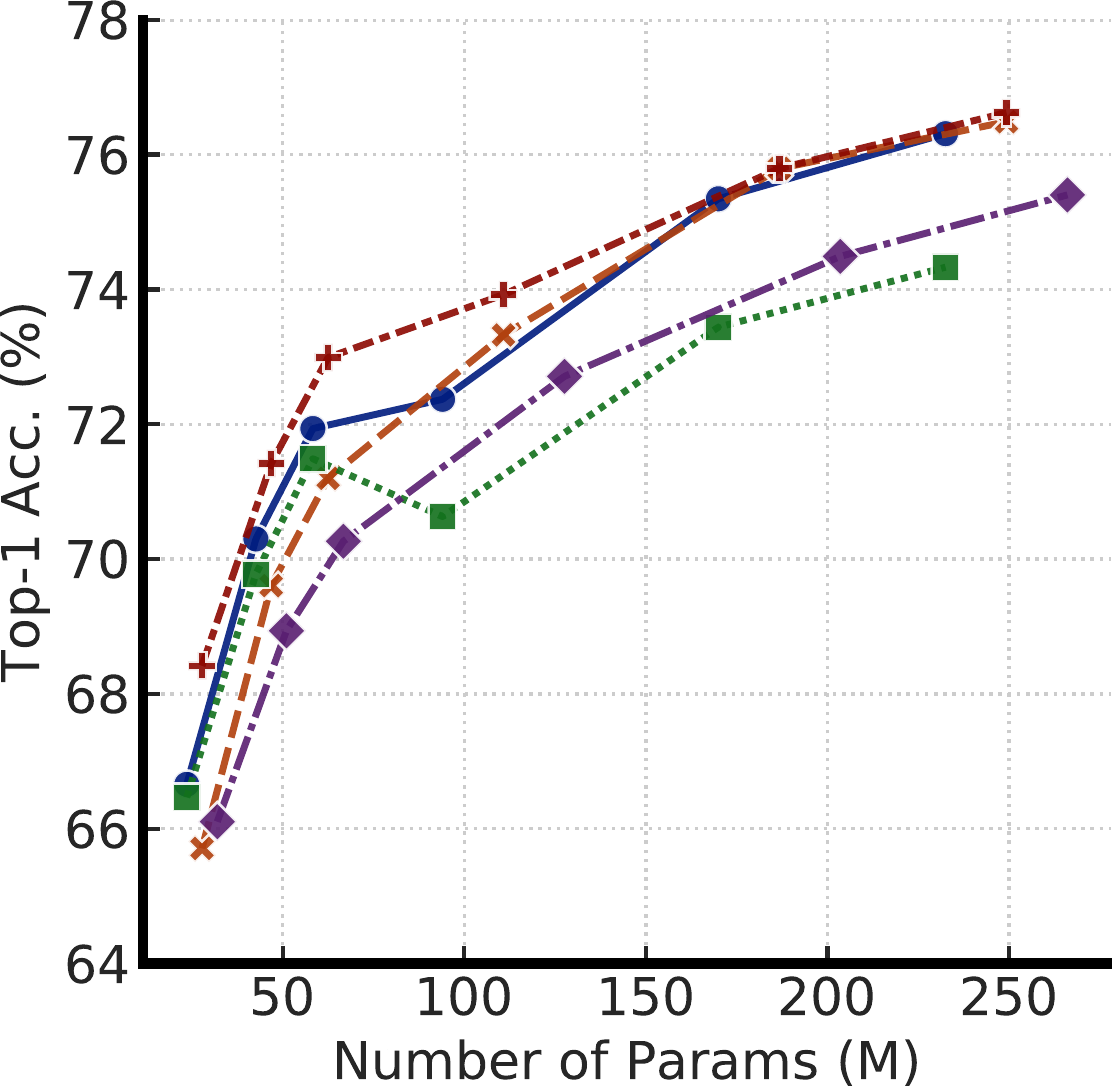}
      \caption{w/o SK (label fraction: 10\%)}
    \end{subfigure}
    \begin{subfigure}{.3\textwidth}
      \centering
      \includegraphics[width=0.98\linewidth]{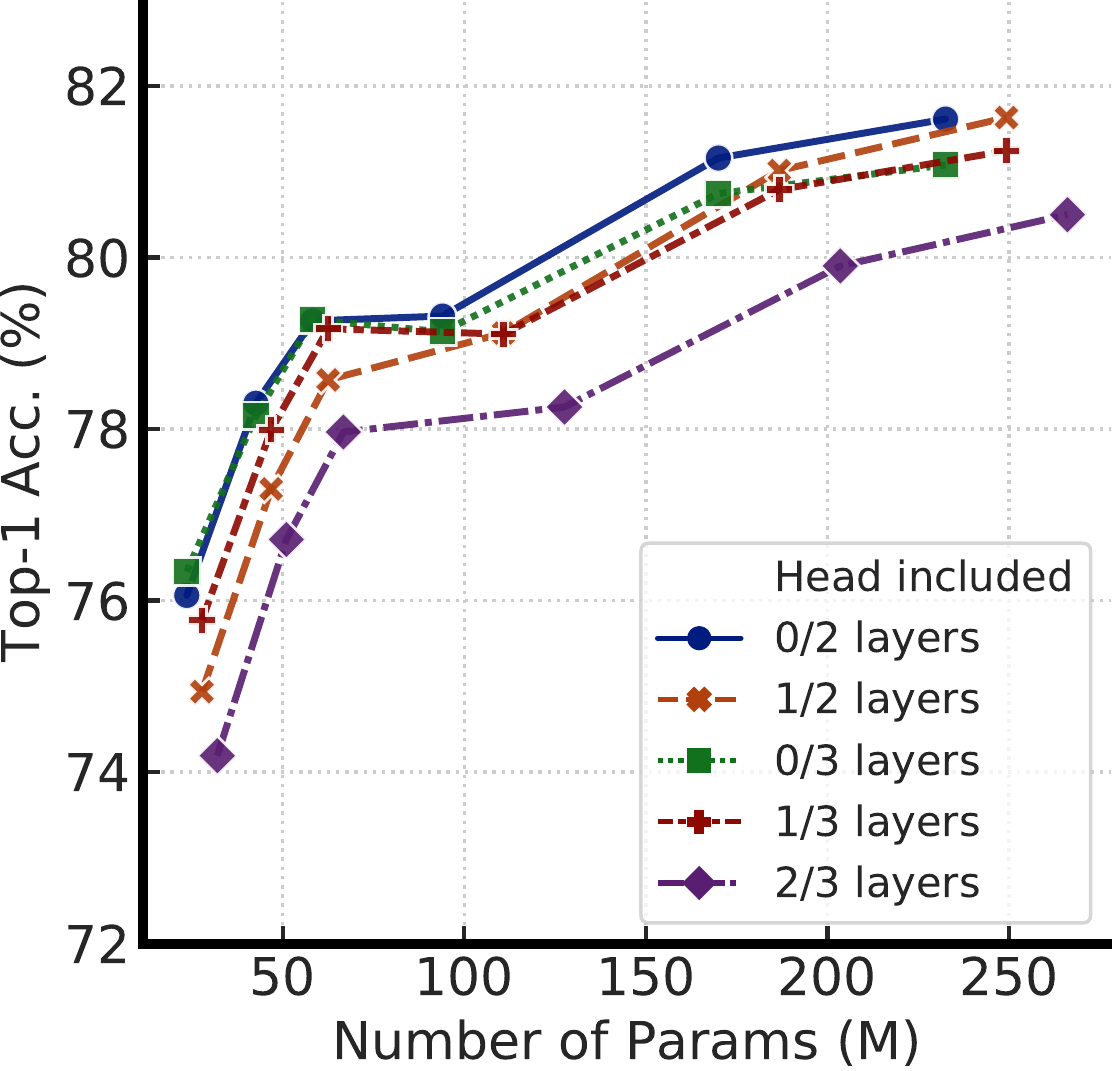}
      \caption{w/o SK (label fraction: 100\%)}
    \end{subfigure}\\
    \begin{subfigure}{.3\textwidth}
      \centering
      \includegraphics[width=0.98\linewidth]{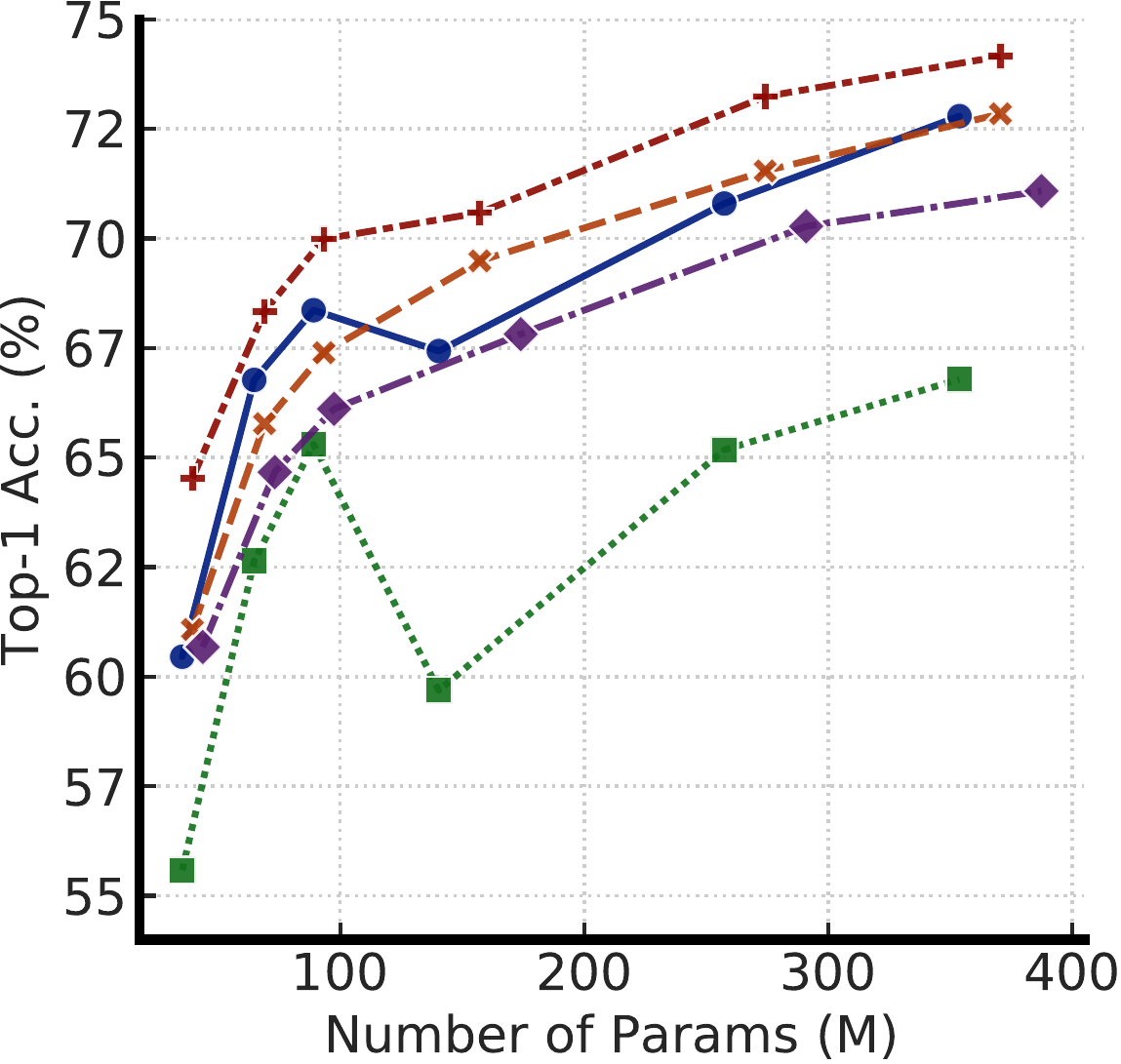}
      \caption{w/ SK (label fraction: 1\%)}
    \end{subfigure}\begin{subfigure}{.3\textwidth}
      \centering
      \includegraphics[width=0.98\linewidth]{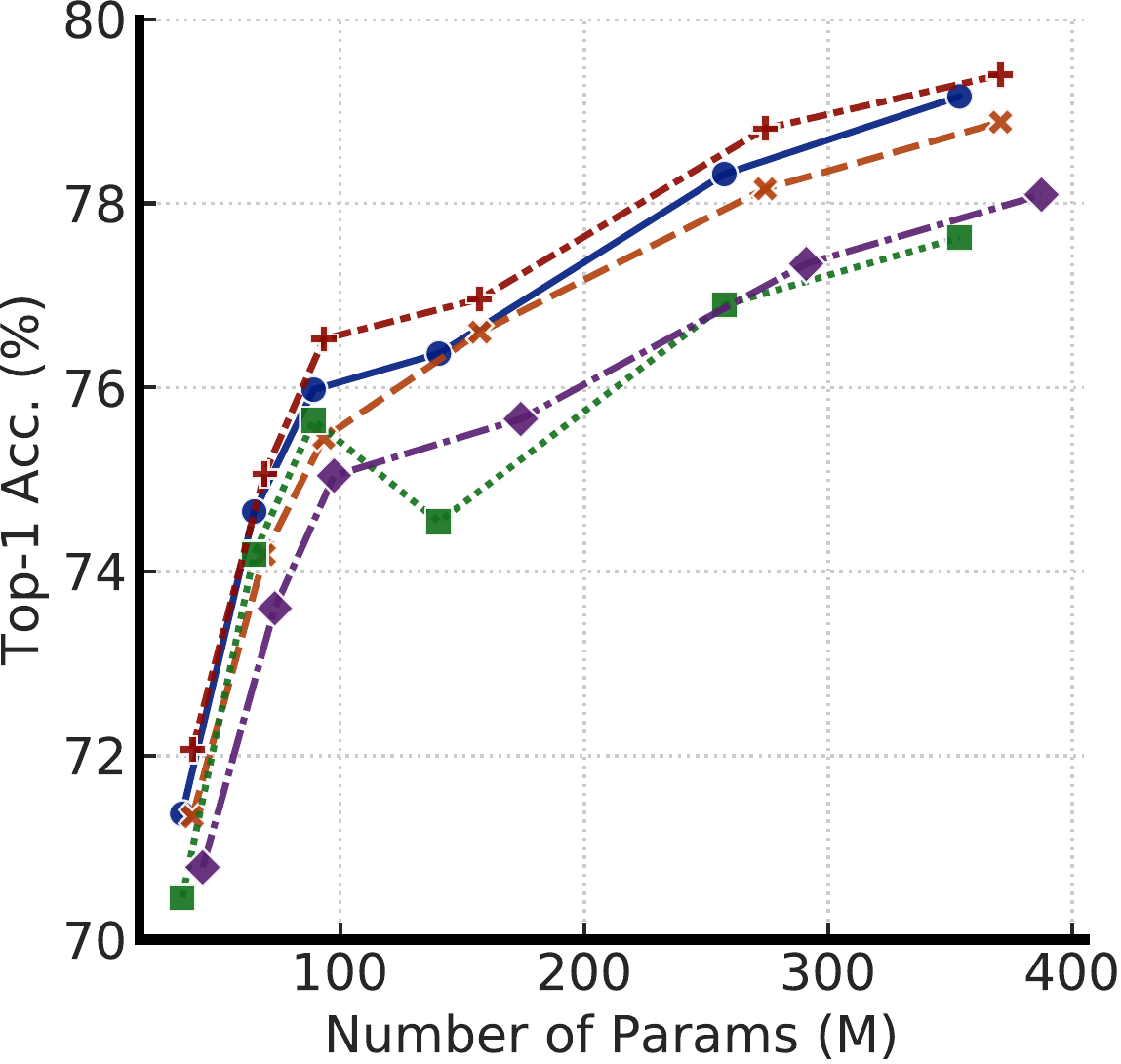}
      \caption{w/ SK (label fraction: 10\%)}
    \end{subfigure}
    \begin{subfigure}{.3\textwidth}
      \centering
      \includegraphics[width=0.98\linewidth]{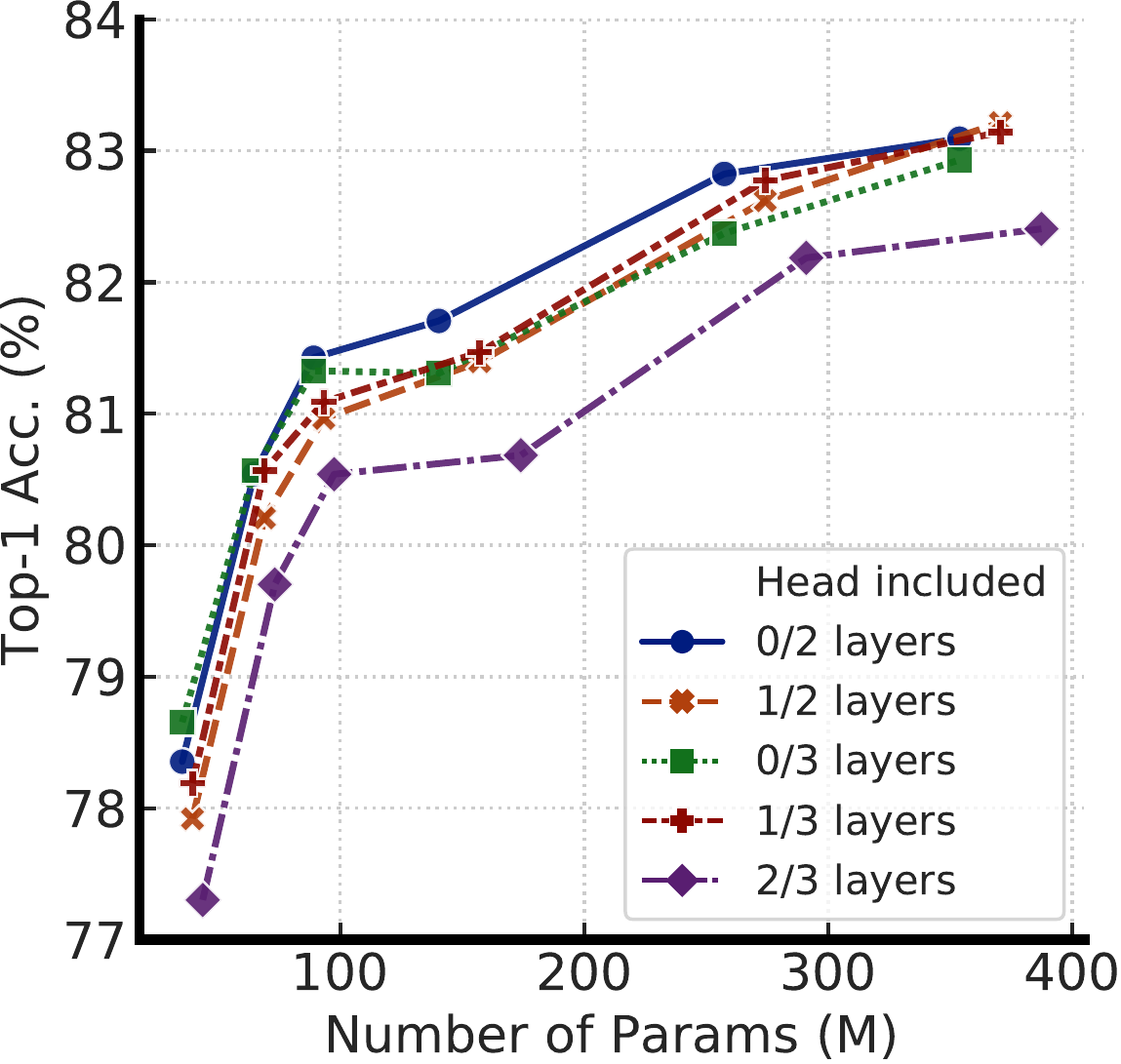}
      \caption{w/ SK (label fraction: 100\%)}
    \end{subfigure}
    \caption{\label{fig:param_head} Top-1 fine-tuning performance under different projection head settings (number of layers included for fine-tuning) and model sizes. With fewer labeled examples, fine-tuning from the first layer of a 3-layer projection head is better, especially when the model is small. Points reflect ResNets with depths of $\{50, 101, 152\}$ and width multipliers of $\{1\times, 2\times\}$. Networks in the first row are without SK, and in the second row are with SK.}
\end{figure}

To understand the effects of projection head settings across model sizes, Figure~\ref{fig:param_head} shows effects of fine-tuning from different layers of 2- and 3-layer projection heads. These results confirm that with only a few labeled examples, pretraining with a deeper projection head and fine-tuning from a middle layer can improve the semi-supervised learning performance. The improvement is larger with a smaller model size.

\begin{figure}[h]
    \centering
    \includegraphics[width=0.8\linewidth]{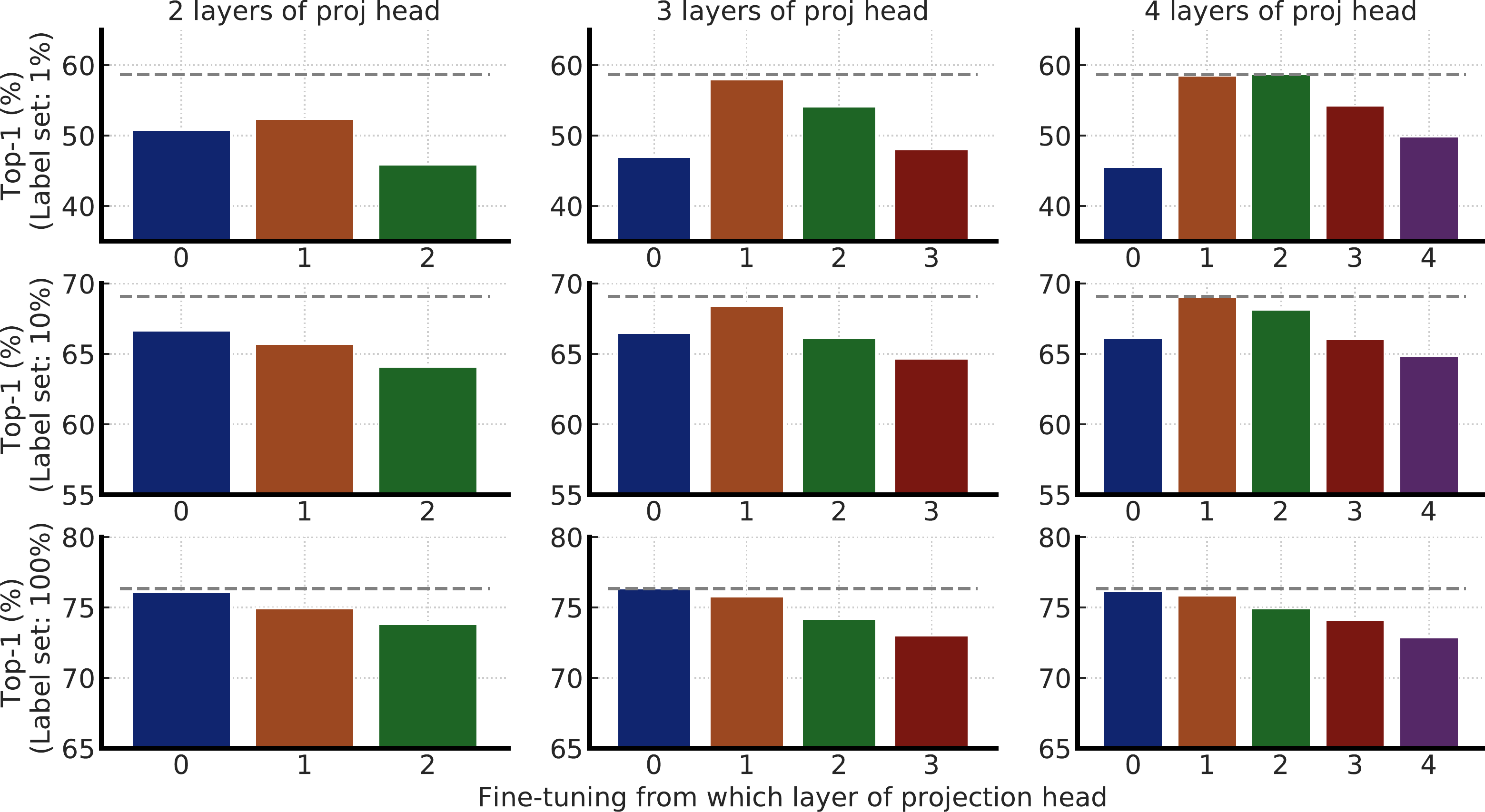}
    \caption{\label{fig:proj_head_study_r50_more} Top-1 accuracy of ResNet-50 with different projection head settings. Deeper projection head help more, when allowing to fine-tune from a middle layer of the projection head.}
\end{figure}

Figure~\ref{fig:proj_head_study_r50_more} shows fine-tuning performance for different projection head settings of a ResNet-50 pretrained using SimCLRv2. Figure~\ref{fig:proj_head_study_r50} in the main text is an aggregation of results from this Figure.

\section{Further Distillation Ablations}
\label{app:distill}

\begin{figure}[h]
    \centering
    \small
    \begin{subfigure}{.4\textwidth}
      \centering
      \includegraphics[width=0.98\linewidth]{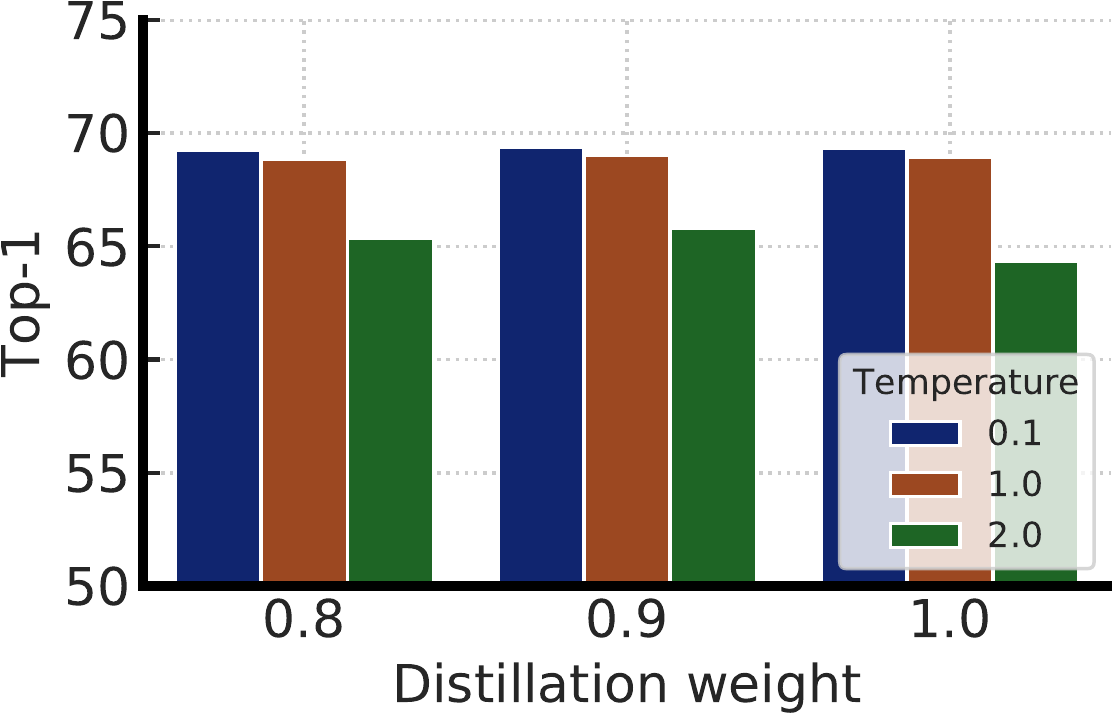}
      \caption{label fraction: 1\%}
    \end{subfigure}\begin{subfigure}{.4\textwidth}
      \centering
      \includegraphics[width=0.98\linewidth]{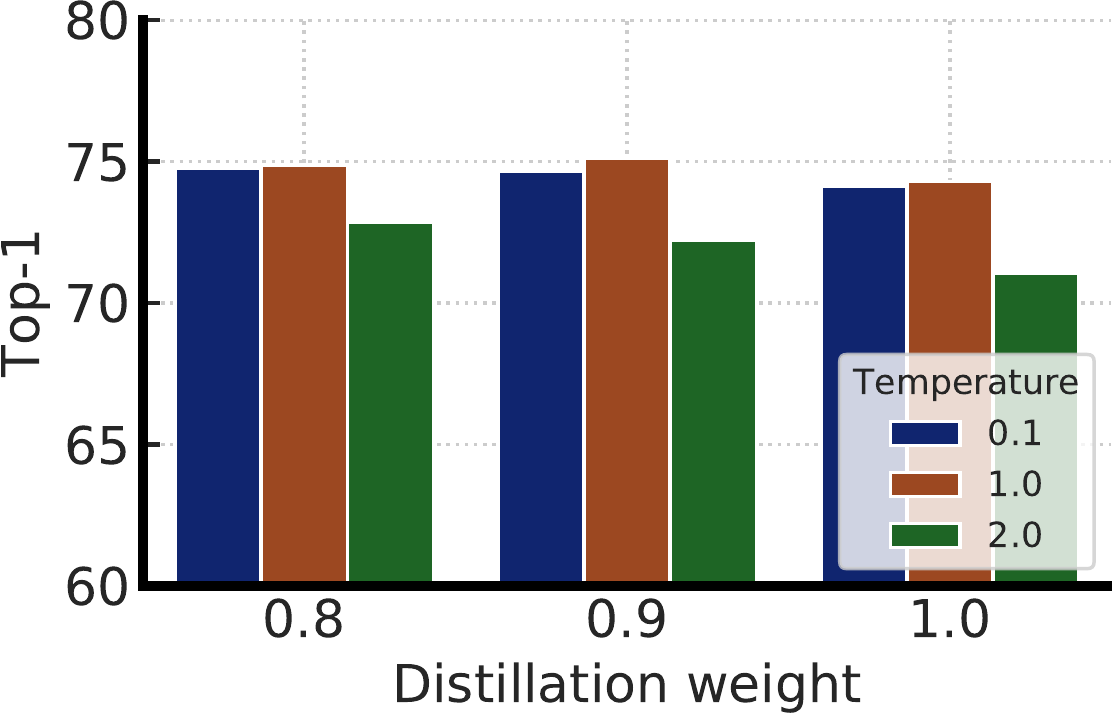}
      \caption{Label fraction: 10\%}
    \end{subfigure}
    \caption{\label{fig:distill_weight} Top-1 accuracy with different distillation weight ($\alpha$), and temperature ($\tau$).}
\end{figure}

Figure~\ref{fig:distill_weight} shows the impact of distillation weight ($\alpha$) in Eq.~\ref{eq:distill_plus}, and temperature used for distillation. We see distillation without actual labels (i.e. distillation weight is 1.0) works on par with distillation with actual labels. Furthermore, temperature of 0.1 and 1.0 work similarly, but 2.0 is significantly worse. For our distillation experiments in this work, we by default use a temperature of 0.1 when the teacher is a fine-tuned model, otherwise 1.0.

\begin{figure}[h]
    \centering
    \small
    \begin{subfigure}{.3\textwidth}
      \centering
      \includegraphics[width=0.98\linewidth]{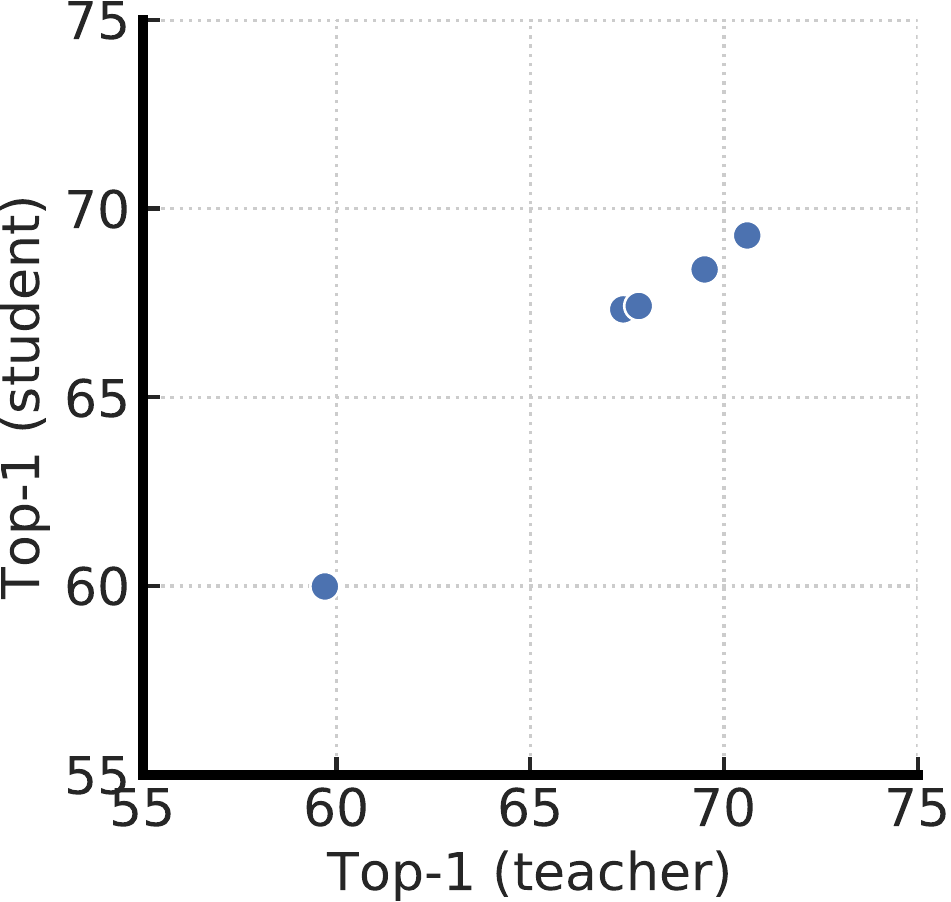}
      \caption{Label fraction: 1\%}
    \end{subfigure}~~~~~~~~~~
    \begin{subfigure}{.3\textwidth}
      \centering
      \includegraphics[width=0.98\linewidth]{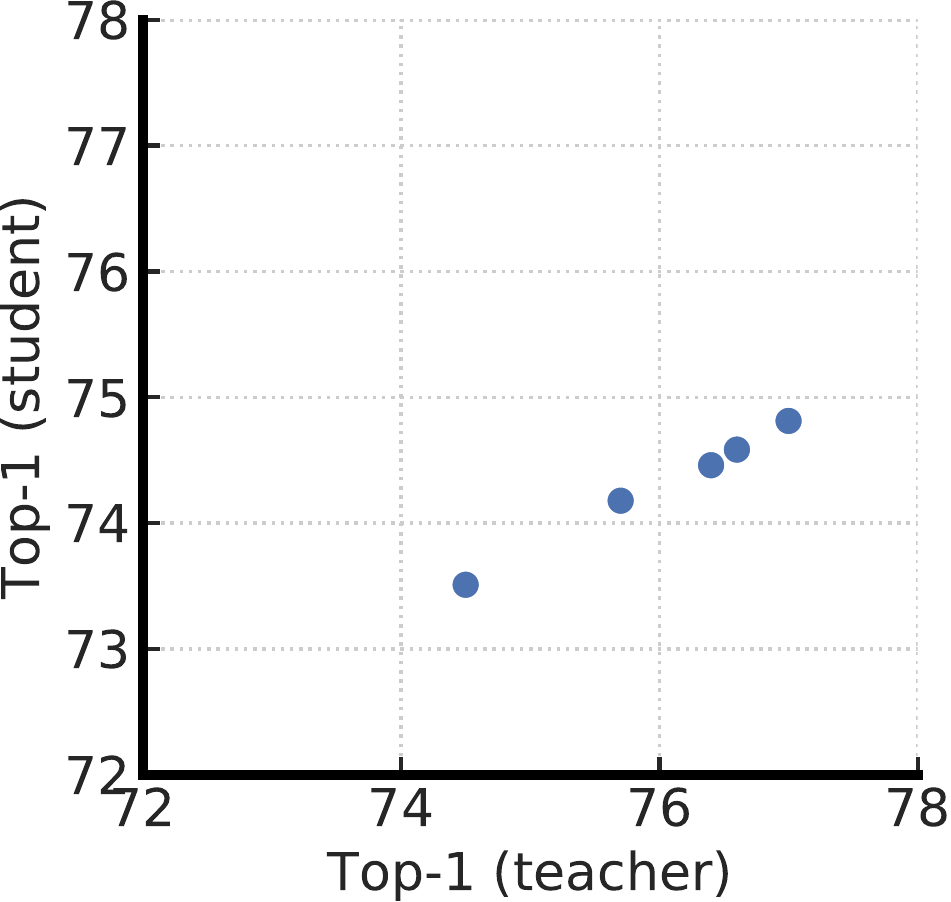}
      \caption{Label fraction: 10\%}
    \end{subfigure}
    \caption{\label{fig:distill_teacher_head} The strong correlation between teacher's task performance and student's task performance.}
\end{figure}

We further study the distillation performance with teachers that are fine-tuned using different projection head settings. More specifically, we pretrain two ResNet-50 (2$\times$+SK) models, with two or three layers of projection head, and fine-tune from a middle layer. This gives us five different teachers, corresponding to different projection head settings. Unsurprisingly, as shown in Figure~\ref{fig:distill_teacher_head}, distillation performance is strongly correlated with the top-1 accuracy of fine-tuned teacher. This suggests that a better fine-tuned model (measured by its top-1 accuracy), regardless their projection head settings, is a better teacher for transferring task specific knowledge to the student using unlabeled data.

\section{CIFAR-10}
\label{app:cifar10}
 
We perform main experiments on ImageNet since it is a large-scale and well studied dataset. 
Here we conduct similar experiments on small-scaled CIFAR-10 dataset to test our findings in ImageNet. 
More specifically, we pretrain ResNets on CIFAR-10 without labels following~\cite{chen2020simple} \footnote{For CIFAR-10, we do not find it beneficial to have a 3-layer projection head, nor using a EMA network. However, we expand augmentations in~\cite{chen2020simple} with a broader set of color operators and Sobel filtering for pretraining. We pretrain it for 800 epochs, with a batch size of 512 and temperature of 0.2.}. The ResNet variants we trained are of 6 depths, namely 18, 34, 50, 101, 152 and 200. To keep experiments tractable, by default we use Selective Kernel, and a width multiplier of 1$\times$. After the models are pretrained, we then fine-tune them (using simple augmentations of random crop and horizontal flipping) on different numbers of labeled examples (250, 4000, and total of 5000 labeled examples), following MixMatch's protocol and running on 5 seeds.

\begin{figure}
\centering
    \includegraphics[width=0.5\textwidth]{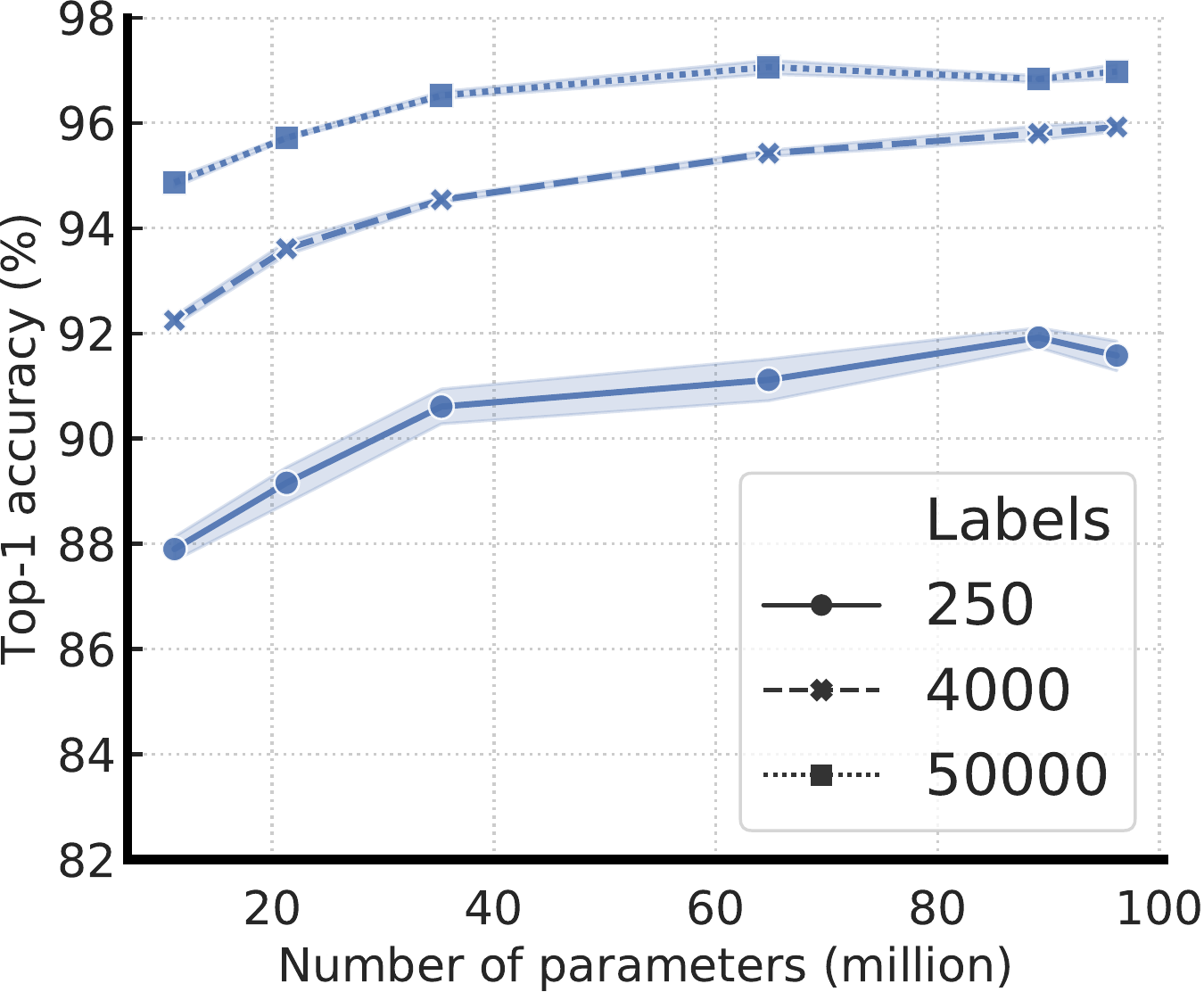}
    \caption{Fine-tuning pre-trained ResNets on CIFAR-10.}
    \label{fig:cifar10_ft}
\end{figure}

The fine-tuning performances are shown in the Figure~\ref{fig:cifar10_ft}, and suggest similar trends to our results on ImageNet: big pretrained models can perform well, often better, with a few labeled examples. These results can be further improved with better augmentations during fine-tuning and an extra distillation step. The linear evaluation also improves with a bigger network. Our best result is 96.4\% obtained from ResNet-101 (+SK) and ResNet-152 (+SK), but it is slightly worse (96.2\%) with ResNet-200 (+SK).

\section{Extra Results}
\label{app:extra_results}

Table~\ref{tab:big_ft_top5} shows top-5 accuracy of the fine-tuned SimCLRv2 (under different model sizes) on ImageNet.

\begin{table}[h!]
\small
\centering
\caption{Top-5 accuracy of fine-tuning SimCLRv2 (on varied label fractions) or training a linear classifier on the ResNet output. The supervised baselines are trained from scratch using all labels in 90 epochs. The parameter count only include ResNet up to final average pooling layer. For fine-tuning results with 1\% and 10\% labeled examples, the models include additional non-linear projection layers, which incurs additional parameter count (4M for $1\times$ models, and 17M for $2\times$ models).}
\label{tab:big_ft_top5}
\begin{tabular}{r@{\hspace{0.35cm}}rr@{\hspace{0.35cm}}r@{\hspace{0.35cm}}r@{\hspace{0.35cm}}r@{\hspace{0.35cm}}r@{\hspace{0.35cm}}r@{\hspace{0.35cm}}r}
\toprule
\multirow{2}{*}{Depth}& \multirow{2}{*}{Width}& \multirow{2}{*}{Use SK~\cite{li2019selective}}& \multirow{2}{*}{Param (M)}& \multicolumn{3}{c}{Fine-tuned on} & \multirow{2}{*}{Linear eval} &  \multirow{2}{*}{Supervised} \\
 &  &  &  &  ~~1\%~ &   ~~10\% &   ~100\% &  &  \\
\midrule
\multirow{4}{*}{50} & \multirow{2}{*}{1$\times$} & False &   \textbf{24} &        \textbf{82.5} &         \textbf{89.2} &          \textbf{93.3} &            \textbf{90.4} &         \textbf{93.3} \\
    &    & True  &   35 &        86.7 &         91.4 &          94.6 &            92.3 &         94.2 \\
    & \multirow{2}{*}{2$\times$} & False &   94 &        87.4 &         91.9 &          94.8 &            92.7 &         93.9 \\
    &    & True  &  140 &        90.2 &         93.7 &          95.9 &            93.9 &         94.5 \\
\multirow{4}{*}{101} & \multirow{2}{*}{1$\times$} & False &   43 &        85.2 &         90.9 &          94.3 &            91.7 &         93.9 \\
    &    & True  &   65 &        89.2 &         93.0 &          95.4 &            93.1 &         94.8 \\
    & \multirow{2}{*}{2$\times$} & False &  170 &        88.9 &         93.2 &          95.6 &            93.4 &         94.4 \\
    &    & True  &  257 &        91.6 &         94.5 &          96.4 &            94.5 &         95.0 \\
\multirow{4}{*}{152} & \multirow{2}{*}{1$\times$} & False &   58 &        86.6 &         91.8 &          94.9 &            92.4 &         94.2 \\
    &    & True  &   89 &        90.0 &         93.7 &          95.9 &            93.6 &         95.0 \\
    & \multirow{2}{*}{2$\times$} & False &  233 &        89.4 &         93.5 &          95.8 &            93.6 &         94.5 \\
    &    & True  &  354 &        92.1 &         94.7 &          96.5 &            94.7 &         95.0 \\
152 & 3$\times$ & True & \textbf{795} & {\textbf{92.3}} & {\textbf{95.0}} & {\textbf{96.6}} & {\textbf{94.9}} & {\textbf{95.1}} \\
\bottomrule
\end{tabular}
\end{table}

\end{document}